\documentclass[10pt,twocolumn,letterpaper]{article}

\usepackage{iccv}
\usepackage{times}
\usepackage{epsfig}
\usepackage{graphicx}
\usepackage{amsmath,amssymb,amsthm,amsfonts,fixmath}  % math stuffs 
\usepackage{nicefrac}       % compact symbols for 1/2, etc.
\usepackage{subcaption}
\usepackage{multirow}

\newcommand{\dis}[1]{\mbox{dist}\left( #1 \right)}

\newcommand{\dpar}[2]{\frac{\partial #1}{\partial #2}}

\DeclareMathOperator*{\argmin}{argmin} % no space, limits underneath in displays
\usepackage{color}  
\definecolor{gray}{RGB}{180,180,180}
\definecolor{lightgray}{RGB}{230,230,230}

\usepackage{algpseudocode,algorithm,algorithmicx}
\algrenewcommand\algorithmicrequire{\textbf{Input:}}
\algrenewcommand\algorithmicensure{\textbf{Output:}}
\theoremstyle{definition}
\newtheorem{definition}{Definition}[section]
 
\theoremstyle{remark}

\theoremstyle{definition}
\newtheorem{proposition}{Proposition}[section]

\usepackage[pagebackref=true,breaklinks=true,letterpaper=true,colorlinks,bookmarks=false]{hyperref}

\iccvfinalcopy % *** Uncomment this line for the final submission

\def\iccvPaperID{****} % *** Enter the ICCV Paper ID here

\ificcvfinal\fi
\begin{document}

% *** My own definitions *** % 
% \theoremstyle{definition}
% \newtheorem{definition}{Definition}[section]
 
% \theoremstyle{remark}
% \newtheorem*{remark}{Remark}

% \theoremstyle{definition}
% \newtheorem{proposition}{Proposition}[section]

%%%%%%%%% TITLE
\title{SalGaze: Personalizing Gaze Estimation using Visual Saliency}

\author{Zhuoqing Chang, Matias Di Martino, Qiang Qiu, Steven Espinosa, and Guillermo Sapiro\\
Duke University, Durham, NC 27708, USA\\
{\tt\small \{zhuoqing.chang, juan.di.martino, qiang.qiu, steven.espinosa, guillermo.sapiro\}@duke.edu}
% \and
% Matias Di Martino\\
% % {\tt\small matias.di.martino@duke.edu}
% \and
% Qiang Qiu\\
% % {\tt\small qiang.qiu@duke.edu}
% \and
% Guillermo Sapiro\\
% % {\tt\small guillermo.sapiro@duke.edu}
% Duke University, Durham, NC 27708, USA\\
}

\maketitle
%\thispagestyle{empty}

%%%%%%%%% ABSTRACT
\begin{abstract}
Traditional gaze estimation methods typically require explicit user calibration to achieve high accuracy. This process is cumbersome and recalibration is often required when there are changes in factors such as illumination and pose. To address this challenge, we introduce SalGaze, a framework that utilizes saliency information in the visual content to transparently adapt the gaze estimation algorithm to the user without explicit user calibration. We design an algorithm to transform a saliency map into a differentiable loss map that can be used for the optimization of CNN-based models. SalGaze is also able to greatly augment standard point calibration data with implicit video saliency calibration data using a unified framework. We show accuracy improvements over 24\% using our technique on existing methods.
\end{abstract}

%%%%%%%%% BODY TEXT
\section{Introduction}
% What is gaze estimation and its applications/importance
Gaze estimation is the problem of estimating a person's line of sight. It is important as eye gaze reflects a person's underlying cognitive process \cite{eckstein2017beyond} which can be used in a wide array of applications including digital content marketing \cite{wedel2008eye}, diagnosing psychiatric conditions such as autism \cite{klin2002visual}, and automated driving \cite{Alletto2016DR(Eye)Ve}.

% What is calibration and why it is needed
Due to differences in the structure and appearance of the eye, calibration is typically required to learn the parameters that are intrinsic to the user in order for gaze estimation algorithms to achieve high precision. This calibration process is typically in the form of having the user look at certain points on the target screen. One main issue with this process is that a one-time calibration typically only works in the same environment setting. Changes in factors such as illumination, head position, and facial appearance can drastically affect the estimation accuracy and recalibrating for each scenario is not feasible. Active gaze calibration can also be too restrictive in some scenarios. For instance, gaze is an important biomarker for autism risk assessment in toddlers \cite{sasson2012eye}, however, it is very challenging to require a toddler to perform active calibration as well as get feedback on the calibration performance. The same challenges also appear for elderly populations with neurodegenerative disorders \cite{anderson2013eye}. Due to these reasons, having a way to passively calibrate for the user is critical for making gaze estimation a more pervasive technology, in particular when using off-the-shelf devices, such as cameras embedded on mobile devices, and deploying in general in-the-wild scenarios.

% Appearance-based gaze estimation and deep learning
With recent advances in machine learning, the combination of appearance-based gaze estimation and deep learning has become a popular method for remote estimation of gaze \cite{xiong2019mixed, chen2019appearance, yu2018deep, wang2019generalizing, fischer2018rt, yu2019improving}. Appearance-based algorithms have the advantage of only using an image of the person's face or eyes as input, therefore eliminating the need of any specialized hardware other than a regular camera. Deep learning algorithms have been shown to be powerful tools for gaze estimation due to its ability to incorporate factors including illumination, head pose, appearance, etc., using end-to-end learning. The biggest challenge with this approach is the need for a huge amount of data with ground truth gaze labels to train the network. Recent works, \cite{CAVE_0324, sugano2014learning, funes2014eyediap, huang2017tabletgaze, zhang2015appearance, krafka2016eye, wood2016learning}, have made great efforts on creating such large datasets, showing promising results. However, the process of collecting labeled gaze data is still a tedious task, and without powerful domain transfer techniques, which have yet to be demonstrated for gaze estimation, such data is limited to the devices, and scenarios for which it was collected. The lack of data as well as an efficient process for collecting data is one of the major bottlenecks of deep learning methods for gaze estimation. Addressing this challenge is the goal of this paper.

% Visual saliency and how it could help
Cognitive science has shown that the human visual system has a strong tendency to focus on highly salient regions in the visual field. There has been extensive work in the area of computational saliency to emulate this behavior. While conventional research in this field leverages ground truth gaze data for visual saliency estimation, in this work we propose to invert the process and utilize visual saliency information for gaze estimation. We argue and demonstrate that saliency information within the scene can be used for calibration purposes without active participation from the user.

% Our work
In this paper, we present SalGaze, a novel framework that leverages visual saliency information, properly processed as here introduced, for personalized gaze estimation using deep learning models. Calibration is transparently performed while the user watches a few short video clips. By using a here proposed differentiable loss map, SalGaze is also able to combine standard point-based calibration data with free-viewing video data under a unified framework, further improving performance if more accuracy is desired. We experiment with both empirical saliency data collected from eye trackers, and saliency data generated from two state-of-the-art saliency algorithms. We show accuracy improvements of more than 24\% using our technique.

%
%-------------------------------------------------------------------------
%------------------------------------------------------------------------
\section{Related work}
\paragraph{Gaze Estimation.}
One of the pioneering works in appearance-based gaze estimation was done by Tan \textit{et al.} \cite{tan2002appearance}. They used over 200 calibration samples to achieve very high accuracy under fixed head pose and illumination setting. Lu \textit{et al.} \cite{lu2011inferring} improved on this by drastically lowering the amount of calibration samples required while maintaining high accuracy. However, their method still required the person's head to be completely still. Sugano \textit{et al.} \cite{sugano2014learning} and Chang \textit{et al.} \cite{chang2016synthesis} used a multi-camera setup and a depth camera respectively to acquire 3D information and synthesize images of the eye under various head poses. An extensive survey of gaze estimation methods can be found in \cite{hansen2009eye}.

Many deep learning based algorithms for gaze estimation have been proposed over the last couple of years. Zhang \textit{et al.} \cite{zhang2015appearance, zhang2017mpiigaze} trained a convolutional neural network (CNN) on over 200,000 images collected from 15 participants over a 3 months period. Wood \textit{et al.} \cite{wood2015rendering, wood2016learning} used advanced computer graphics to synthesize large amounts of eye images to train CNN models. Shrivastava \textit{et al.} \cite{shrivastava2017learning} used a generative adversarial network (GAN) to produce more realistic synthesized eye images. Krafka \textit{et al.} \cite{krafka2016eye} used Amazon Turk to collect over 2 million images of people using iPhones and iPads. Park \textit{et al.} \cite{park2018learning, park2018deep} trained a deep network to regress to intermediate eye landmarks and pictorial representations of the eyeball before estimating the gaze. These contributions have made significant breakthroughs in the area of calibration-free person-independent gaze estimation which shows the importance of having large amounts of data. Our work aims to extend these methods by allowing the use of personalized data that can be collected at scale.

\paragraph{Gaze Personalization.}
\label{subsection:gaze_personalization}
Appearance-based algorithms have mostly been focused on person-independent gaze estimation. To the best of our knowledge, only a couple of works have tried to address the challenge of personalizing the algorithm to a specific user. Krafka \textit{et al.} \cite{krafka2016eye} used the CNN features of a few calibration samples to train a Support Vector Regression model. Several methods \cite{linden2018appearance, chen2019appearance, xiong2019mixed} have tried to explicitly incorporate person-dependent parameters in the model. These parameters are estimated with a few samples during testing. Others \cite{yu2019improving, sugano2012appearance} use synthesis techniques to augment the number of samples for a specific user to train a person-specific model or fine-tune a generic model. The above methods all require explicit calibration to collect limited samples of the user. Due to the transparent nature of our data collection process, we are not limited to collecting only a few samples from a new user, therefore allowing for better personalization.

\paragraph{Saliency Prediction.}
\label{subsection:related_work_saliency}
Seminal work in this area was done by Itti \textit{et al.} \cite{itti1998model}. They presented a computational model that extracted low-level features such as color and orientation to predict a global saliency map. More recently, a plethora of deep learning based methods have been proposed for static saliency prediction. K{\"u}mmerer \textit{et al.} \cite{kummerer2014deep, kummerer2016deepgaze} proposed two deep saliency prediction networks, DeepGaze I and DeepGaze II, that was built on the AlexNet \cite{krizhevsky2012imagenet} and VGG-19 \cite{simonyan2014very} models respectively. Pan \textit{et al.} \cite{pan2017salgan} used a GAN to generate saliency maps. Cornia \textit{et al.} \cite{cornia2018predicting} and Liu and Han \cite{liu2018deep} combines Long Short-Term Memory networks (LSTM) \cite{hochreiter1997long} with ResNet \cite{he2016deep} to infer saliency by incorporating global and scene contexts. The problem of saliency prediction is not the focus of this work, we use saliency as a tool for gaze estimation. Further progress on computational saliency estimation can potentially further improve the results we obtain in this paper.

\paragraph{Gaze and Saliency.}
While the problem of gaze estimation and saliency prediction have received a lot of attention in each of their own respective areas, there are only a handful of works, to date, that make a connection between them. Sugano \textit{et al.} \cite{sugano2010calibration} were the first to utilize saliency as a probability map for gaze estimation. They use Gaussian process regression to establish a mapping between eye images and gaze positions on a monitor under a fixed head pose setting. Chen \textit{et al.} \cite{chen2011probabilistic} used a model-based approach where the parameters of the eyeball are estimated in a probabilistic manner using  saliency information of the stimuli. Their method is based on the pupil center corneal reflection (PCCR) which requires the use of an infrared camera in order to locate the pupil position. Many other works \cite{recasens2015they, recasens2017following, fan2018inferring, soo2015social, chong2018connecting} uses the saliency information of an image to determine if a person in the same image is looking at a salient object. Our work differs from them as we use the saliency information of an out-of-frame target for precise gaze estimation. Contrary to these works, our method is designed for deep learning algorithms that utilize a loss function, and as such we design a new saliency-informed differentiable cost function which is also capable of combining saliency information with point-wise calibration. The ability of transparently collecting large amounts of gaze data using saliency information further enhances the potential of our method.

%-------------------------------------------------------------------------
\section{Personalized Gaze Estimation from Saliency}
%-------------------------------------------------------------------------
In Section \ref{subsection:point_loss}, we mathematically formulate the problem of 2D gaze estimation for standard point-wise calibration data. Then in Section \ref{subsection:map_loss}, we extend the formulation to using saliency information and derive a solution by designing a differentiable loss map. We show that traditional point-based calibration is a special case of our solution and therefore can be combined with it. Implementation details of our CNN model are provided in Section \ref{subsection:implementation}.

\subsection{Point Loss}
\label{subsection:point_loss}
Let $\Omega\subset\mathbb{R}^2$ denote an open set where a person's gaze is tracked, e.g. a computer monitor or a phone screen, $I$ the input, typically the face or eye image of the person or a combination of them, and $f$ a model capable of estimating the person's gaze $\hat{p}$ from this image, i.e., $\hat{p} = f(I)$ ($\hat{p}\in\Omega$). 

The standard calibration procedure consists of collecting images $I_1, ..., I_n$ of users looking at pre-specified locations $p_1, ..., p_n$ ($p_i\in\Omega$) on the screen. The parameters of the model $f$ are denoted as ${\theta}=(\theta_1, ..., \theta_m)$. We can optimize these parameters such that the empirical error on the collected data is minimized,
\begin{equation}\label{eq:standard_l2_optimization}
    \theta_{opt} = \argmin_{\theta} \sum_{i=1}^n d(f_\theta(I_i),p_i)^2 \, ,
\end{equation}

\noindent where $d$ represents some distance between the predicted gaze $\hat{p}=f(I)$ and the ground truth gaze $p$, e.g., the one induced by the L2 norm $d(u,v) = \|u-v\|_2 = \sqrt{\sum_k (u_k-v_k)^2}$. Equation \eqref{eq:standard_l2_optimization} can be solved, for example, using stochastic gradient descent, as the loss $\mathcal{L}(\theta)= \sum_{i=1}^n d(f_\theta(I_i),p_i)^2$ is differentiable, 
\begin{equation}
   \dpar{\mathcal{L}_i}{\theta_j} = 2(f_{\theta}(I_i)-p_i)^T \dpar{f_{\theta}(I_i)}{\theta_j}.
\end{equation}

Since $f$ is implemented using differentiable CNN models in this work, $\dpar{f_{\theta}}{\theta}$ is well defined and can be computed numerically. Gradient descent leads to the updating rule, 
\begin{equation}\label{eq:grad_descent}
    \theta^{t+1} = \theta^{t} - \delta\, (\underbrace{f_\theta(I)}_{\hat{p}}-p)^T \dpar{f_{\theta}(I)}{\theta} \, ,
\end{equation}

\noindent where $\delta$ represents the gradient descent step size. Essentially, the parameters of the model are slowly modified such that $\hat{p}\rightarrow p$. As shown in Eq.~\eqref{eq:grad_descent}, the term proportional to $-(\hat{p}-p)$ pushes the prediction of the model in the opposite direction of the error vector, therefore improving the accuracy. This simple optimization technique has been shown to be very robust in the context of gaze estimation \cite{krafka2016eye}.

\subsection{Probability Map Loss}
\label{subsection:map_loss}
We extend the ideas in the previous section to the scenario where, instead of precise point calibration data, we have, for each input image $I_i$, saliency information $s_i:\Omega\rightarrow[0,1]$ of the content the user is watching. $s(x,y)$ can be interpreted as a measure of the likelihood that the user is looking at the point $(x,y)\in\Omega$.

A naive way to adapt Eq. \eqref{eq:standard_l2_optimization} to exploit the new calibration data $\{s_i\}$ would be, 
\begin{equation}\label{eq:naive_implementation}
    \theta_{opt} = \argmin_{\theta} \sum_{i=1}^n g(s_i(f_\theta(I_i))),
\end{equation}

\noindent where $g$ is defined as some smooth monotonic decreasing function, e.g., $g(u)=-u$ or $g(u) = 1/u$. This formulation makes sense in regions where $s(x,y) \neq 0$ since if the model predicts a position $f_\theta(I)$ of high probability, $s(f_\theta(I))$ would be large, and the loss $g(s(f_\theta(I)))$ would be small. However, for regions where $s(x,y) = 0$, which could often occur in saliency maps, $\|\nabla s\|$ is zero and a gradient descent-like optimization technique will fail.

In order to obtain a well-posed and robust optimization scheme, we propose to compute a loss map $l(s)$ with the following properties: 
\begin{itemize}
  \item $l(s)$ should be continuous, differentiable and have properties described in \ref{prop:prop_distance};
  \item it should encourage predictions to occur at regions with large saliency values;
  \item for point data, $l(s)$ should represent the distance to this point as in Eq. \eqref{eq:standard_l2_optimization}.
\end{itemize}
To that end, we adapt ideas from \cite{Aubert2010} and implement a \emph{Reinitialization-like} equation as we detail next.

First we set a threshold $\lambda$ and compute a binary map $l_0:\Omega\rightarrow\{0,1\},\ l_0(x,y) = 1$ if $s(x,y) < \lambda$ and $0$ otherwise. Then we use this binary image as the initial condition of the Partial Differential Equation (PDE)
\begin{equation}\label{eq:reinitialization_pde}
    \left\{ 
    \begin{array}{l}
    \displaystyle\dpar{u((x,y),t)}{t} + l_0(x,y)(\|\nabla u((x,y),t)\| - 1) = 0, \vspace{2mm}\\
    u((x,y),0) = l_0(x,y).
    \end{array} \right.
\end{equation}

\begin{figure*}[h]
\centering

\begin{subfigure}{0.246\textwidth}
\includegraphics[width=1\linewidth]{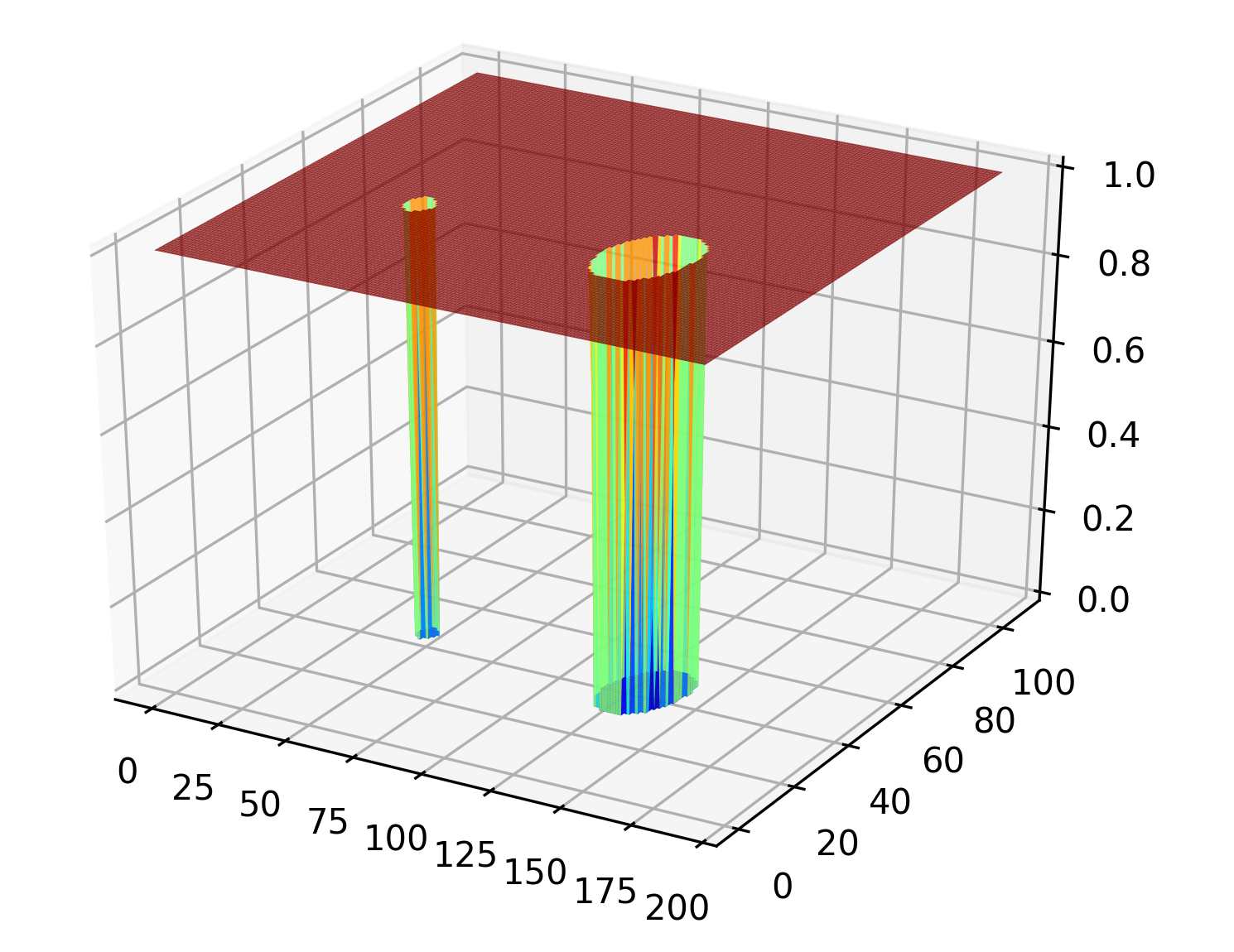}
\caption{Iteration 0}
\end{subfigure}
\begin{subfigure}{0.246\textwidth}
\includegraphics[width=1\linewidth]{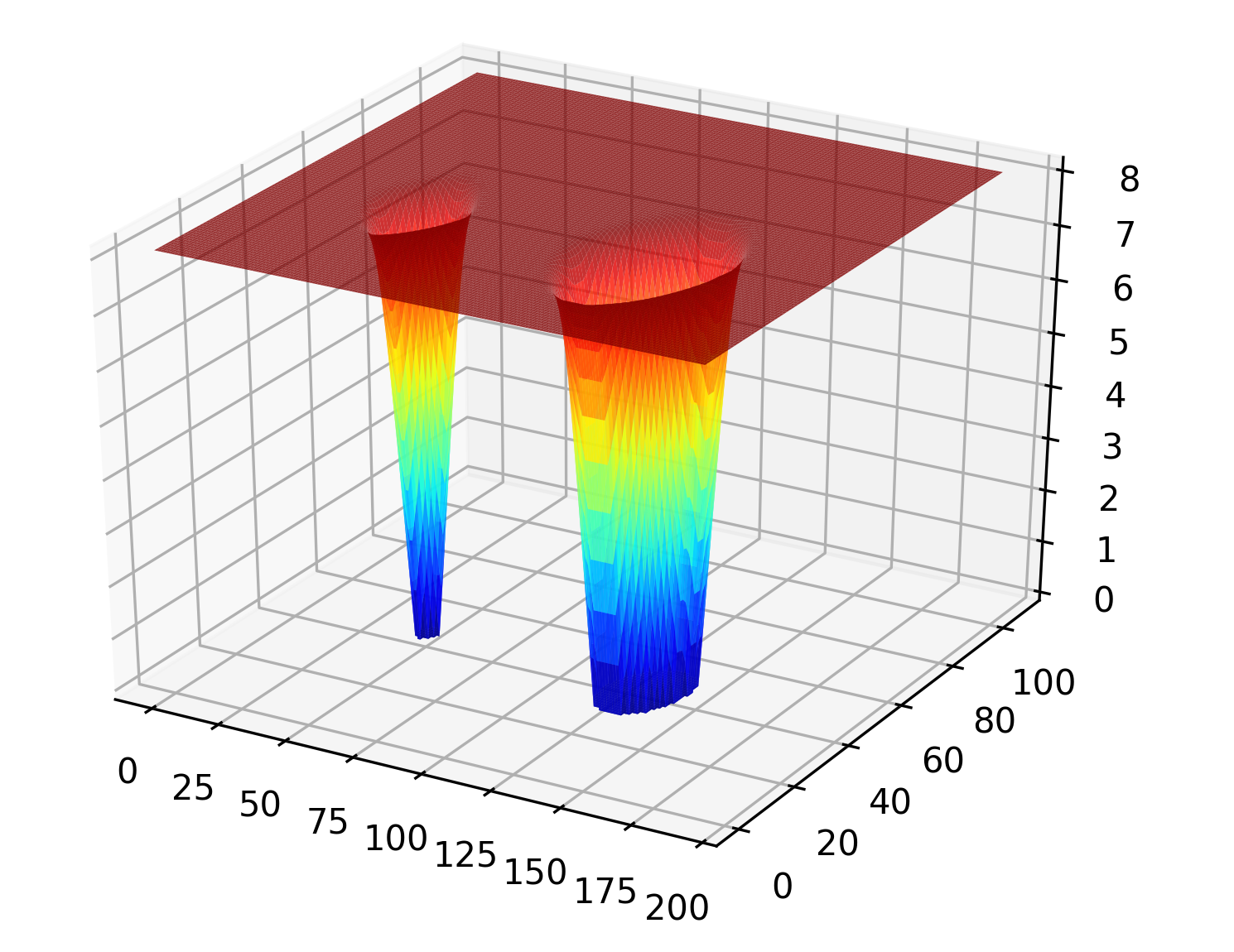}
\caption{Iteration 100}
\end{subfigure}
\begin{subfigure}{0.246\textwidth}
\includegraphics[width=1\linewidth]{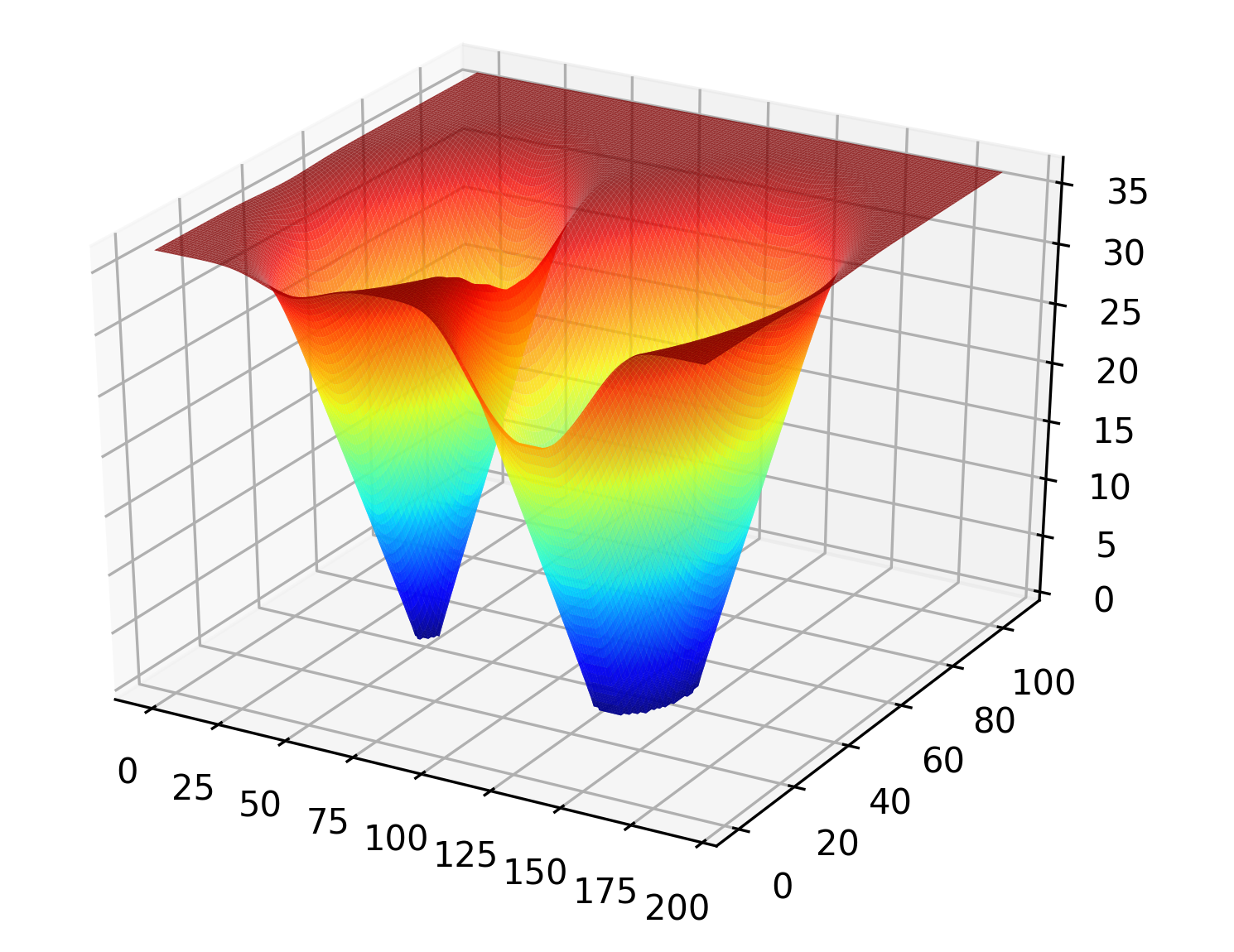}
\caption{Iteration 500}
\end{subfigure}
\begin{subfigure}{0.246\textwidth}
\includegraphics[width=1\linewidth]{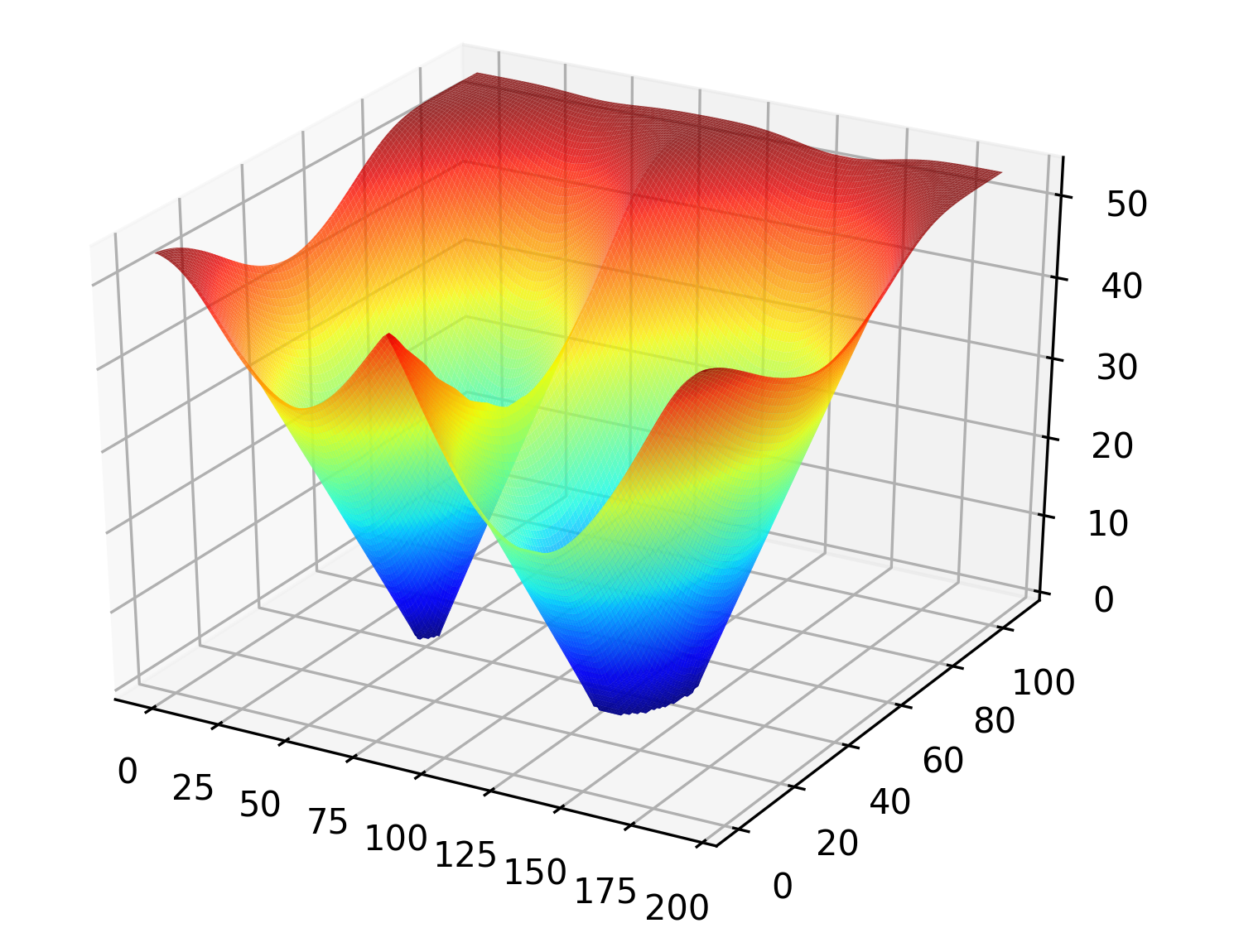}
\caption{Iteration 1000}
\end{subfigure}

\caption{Evolution from binary map to loss map. (a), (b), (c), and (d) shows the loss map at different iterations of the reinitialization algorithm.}
\label{fig:reinitialization}
\end{figure*}

Figure \ref{fig:reinitialization} illustrates the evolution of $u((x,y),t)$. Algorithm \ref{alg:reinitialization} describes the numerical implementation of the reinitialization scheme described in Eq.~\eqref{eq:reinitialization_pde}. A robust implementation of the gradient computation step in Algorithm \ref{alg:reinitialization} is given in Algorithm \ref{alg:gradient}.

\begin{definition}
Let $\Gamma$ be a close set in $\Omega$ and $\dis{p,\Gamma}$ the distance of a point $p$ to the set $\Gamma$ defined as 
\begin{equation}
    \dis{p,\Gamma} = \inf_{q\in\Gamma}\|q-p\|_2.
\end{equation}
\end{definition}

\begin{definition}
The skeleton of $\Gamma$, denoted by $S_\Gamma$, is the set of points $x\in\mathbb{R}^2$ such that there exist at least two distinct points $y$ and $z$ in $\Gamma$ satisfying
$$
|x-y|=|x-z|=\dis{x,\Gamma}.
$$
\end{definition}

\begin{proposition}\label{prop:prop_distance}
If $\Gamma$ is a closed subset of $\Omega$, $u(p)=\dis{p, \Gamma}$, and $\bar{S_\Gamma}$ denotes the closure of the skeleton of $\Gamma$,
\begin{enumerate}
    \item $u$ is 1-Lipschitz for all $p\in\Omega$,
    \item $\|\nabla u\| = 1$ for all $p\in\Omega\backslash\bar{S_\Gamma}$.
\end{enumerate}
\end{proposition}

\begin{proposition}\label{prop:solution}
Let $\Gamma=\{q\in\Omega/ l_0(q) = 0\}$. The function $u: \Omega\times\mathbb{R}_+\rightarrow\mathbb{R}$ defined by, 
\begin{equation}\label{eq:solution_reinit}
    u(p, t) = \left\{ 
    \begin{array}{ll}
    \displaystyle \mbox{inf}_{|q|\le t} (l_0(p+q) + t) & \mbox{if}\ t\le t_p, \vspace{2mm}\\
    \dis{p,\Gamma} & \mbox{if}\ t>t_p,
    \end{array} \right.
\end{equation}
where 
$$
t_p = \mbox{inf}\left\{t\in\mathbb{R}_+ / \mbox{inf}_{|q|\le t} (l_0(p+q)) = 0 \right\} 
$$
is the unique solution of \eqref{eq:reinitialization_pde} uniformly continuous on $\Omega\times[0, T]$, $\forall\, T>0$ and vanishing on $\Gamma\ \ \forall t\in[0, T]$. 
\end{proposition}

Proposition~\ref{prop:prop_distance} is similar to the Proposition 4.3.1 presented in \cite{Aubert2010} in that we define a positive distance function instead of a signed distance function, and $\Gamma$ as a closed subset of $\Omega$ instead of a closed curve in $\mathbb{R}^2$. In addition, Proposition~\ref{prop:solution} can be proved with a straightforward adaptation of the proof of Theorem 4.3.4 presented in \cite{Aubert2010}.

We define $l(x,y)$ as the solution of Eq.~\eqref{eq:reinitialization_pde} with initial condition $l_{0}(x,y)$, i.e., $l(x,y) = u((x,y),t^*)$ for any $t^*$ larger than the longest straight line contained in $\Omega$, therefore, $l(x,y)$ is the distance of $(x,y)$ to the set defined by $\{q\in\Omega\ /\ l_0(p)=0\}$. Finally, given the data $\{I_i, s_i\}$, we compute the corresponding loss maps $\{l_i\}$, and adapt the gaze model by, 
% ($s_i\rightarrow l_{0i} \rightarrow l_{i}$)
\begin{equation}\label{eq:out_implementation}
    \theta_{opt} = \argmin_{\theta} \sum_{i=1}^n l_{i}(f_\theta(I_i))^w.
\end{equation}

We can chose $w=1$ or $w=2$ to minimize the distance or the squared distance respectively. We refer to the former as the \emph{map loss} and the latter as the \emph{squared map loss}. When the prediction $\hat{p}$ is outside the valid set $l_0=0$, the gradient of our defined loss will update the model to predict towards the closest point in the set. The norm of the gradient is proportional to the distance to the set when $w=2$, or a unitary vector when $w=1$. As we will show in Section \ref{section:calibration}, $w=2$ leads to a faster convergence rate, while $w=1$ is more robust to outliers.

It is important to mention that point calibration data can be seen as a special case where for a given calibration point $p$, its equivalent saliency map is $s(x,y) = 0,\ \forall (x,y)\in\Omega\backslash\{p\}$ and $s(p) = 1$. Therefore our formulation allows the use of both point and saliency data sources for gaze estimation.

% and represent some distance between the ground truth and the predicted gaze when point calibration information is provided (i.e., when $s_i(x,y) = 0\ \forall (x,y)\in\Omega\backslash\{p_i\}$ and $s_i(p_i) = 1$).

\begin{algorithm}
	\caption{Reinitialization Equation. Implementation of the solution of the PDE described in Eq.~\eqref{eq:reinitialization_pde}.}
	\label{alg:reinitialization}
	\begin{algorithmic}[1]
		\Require{binary loss map $l_0: [1 ... H]\times [1...W] \rightarrow \{0, 1\}$}
		\State Set iteration parameters:
        \State tol $= 0.01$ stop criteria
		\State $\delta = .1$ step size
		\State Initialize variables.
		\State diff $=$ tol $+ 1$
		\State $u_0 = l_0$
		\While {diff $>$ tol}
		    \State $u_{0x},\ u_{0y}\ =$ computeGradient($u_0$)  (see Alg.~\ref{alg:gradient})
		    \State $u_1 = u0 - \delta \, l_0 \left( \sqrt{ u_{0x}^2 + u_{0y}^2}-1\right)$
		    \State diff $= \|u_1 - u_0\|_2$
		    \State $u_0=u_1$
		\EndWhile
		\State \Return{$u_1$} 
	\end{algorithmic}
\end{algorithm}

\begin{algorithm}
	\caption{computeGradient (robust implementation).}
	\label{alg:gradient}
	\begin{algorithmic}[1]
		\Require{$u: [1...H] \times [1...W] \rightarrow [0, 1]$.}
		\State Initialize 
		\State $\delta_x^+ = zeros(H,W)$
		\State $\delta_y^+ = zeros(H,W)$
		\State $\delta_x^- = zeros(H,W)$
		\State $\delta_y^- = zeros(H,W)$
		\State Compute the nonoscillatory upwind scheme (see \cite{Aubert2010} appendix A.3 for details).
		\State $\delta_x^+[:,2:-2] = \mbox{max}(u[:,2:-2]-u[:,1:-3],0)$
		\State $\delta_y^+[2:-2,:] = \mbox{max}(u[2:-2,:]-u[1:-3,:],0)$
		\State $\delta_x^-[:,2:-2] = -\mbox{min}(u[:,3:-1]-u[:,2:-2],0)$
		\State $\delta_x^-[:,2:-2] = -\mbox{min}(u[3:-1,:]-u[2:-2,:],0)$
		\State 
		\State $u_x = \mbox{max}(\delta_x^+, \delta_x^-)$
		\State $u_y = \mbox{max}(\delta_y^+, \delta_y^-)$
		\State \Return{$u_x,\ u_y$} 
	\end{algorithmic}
\end{algorithm}

%------------------------------------------------------------------------
\subsection{Implementation Details} \label{subsection:implementation}
%------------------------------------------------------------------------
% Gaze model -------------------------------------------------------------
Our model architecture is based on iTracker \cite{krafka2016eye}, a CNN that predicts the user's gaze position on an iPhone or iPad. The network uses images of the user's face collected from the front facing camera of the device as input, and outputs the position on the screen of where the user is looking at. Specifically, the face and two eye regions of the image as well as the face grid, a binary matrix representing the spacial position of the face inside the image, are used as 4 inputs to the network. The output is the 2D gaze position relative to the camera. This technique allows for pooling data from different device models and device orientations to train a relatively robust model, although its performance still degrades when extrapolating to new conditions (e.g., different devices or acquisition scenarios).

\begin{figure}[h]
\centering
\includegraphics[width=1\linewidth]{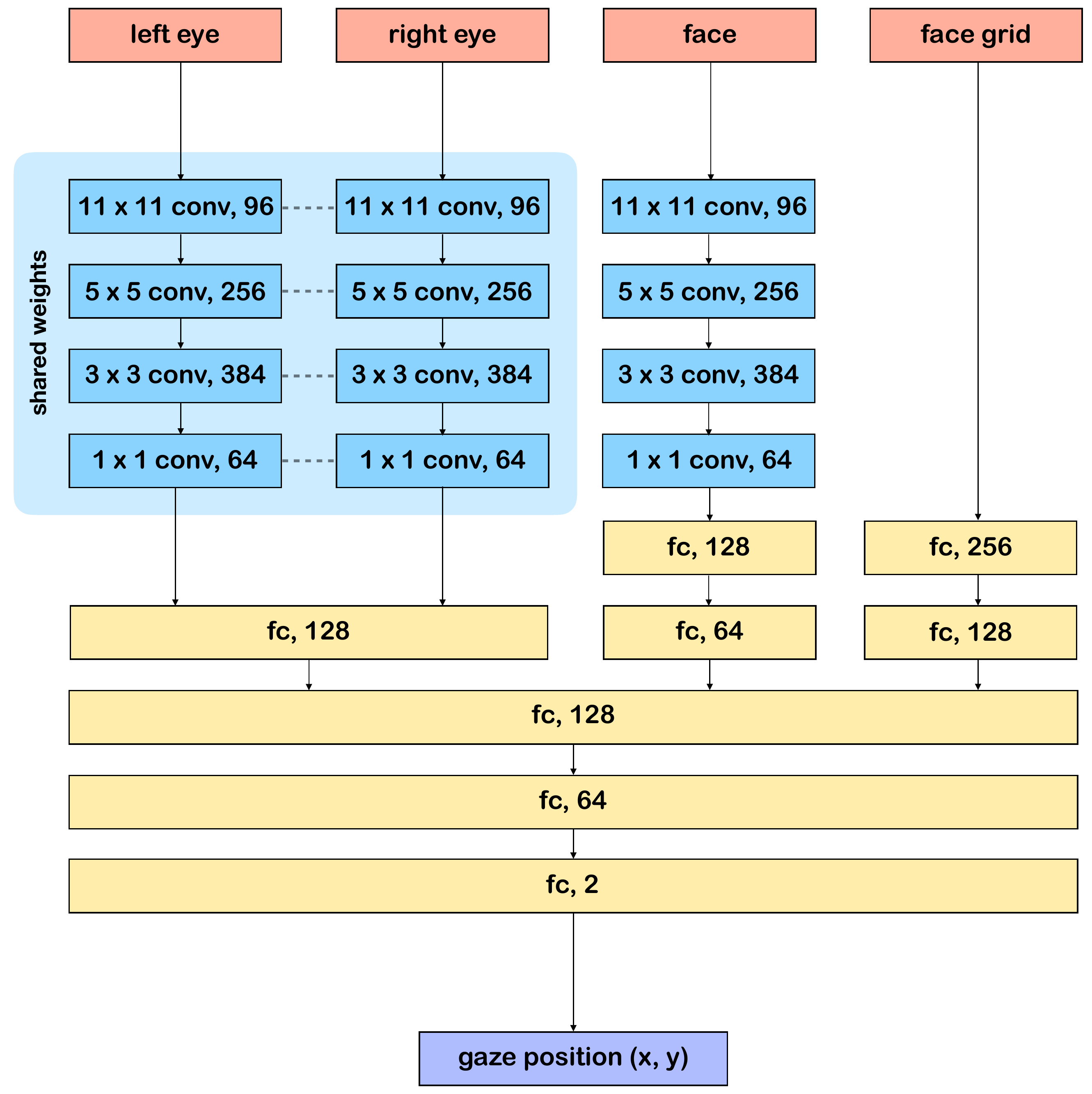} 
\caption{Network architecture}
\label{fig:model_architecture}

\end{figure}

We modify the iTracker architecture in the following way: (1) The size of the input face and eye images is reduced from 224 $\times$ 224 to 64 $\times$ 64 to reduce training time. (2) An extra fully connected layer is added at the end of the network for fine-tuning purposes. (3) The local response normalization layer after each convolutional block is replaced with the superior batch normalization layer. (4) The mean squared loss is replaced with our custom map loss defined in Section \ref{subsection:map_loss}. The final model architecture is shown in Figure \ref{fig:model_architecture}. Adapting a known architecture to the new personalized formulation helps to illustrate the plug-and-play style of our proposed framework.

Our model is implemented in Python using Tensorflow. The optimization is performed using the Adam \cite{kingma2014adam}.
%------------------------------------------------------------------------
\section{Experiments}
%------------------------------------------------------------------------
In this section, we perform experiments to evaluate how different hyperparameters, data quantity, and different types of saliency computations affect the performance of our method.

\subsection{Data Collection}
We asked 9 participants (7 male, 2 female) to each record 10 sessions of data using an iPad Air 2 in landscape orientation. Each session is composed of 2 phases: a point phase and a video phase. The point phase is similar to a traditional calibration procedure where 20 points sequentially appear at random locations on the screen for 2 seconds each. In the video phase, several short 15 to 20 second video clips are played with rest intervals between them. The 10 sessions of data for each participant are collected in different locations across several days to encourage variability in pose, illumination, and general acquisition environments.

The videos we show to participants come from two public saliency databases: SAVAM \cite{lyudvichenko2017semiautomatic} and Coutrot Database 1 \cite{coutrot2013toward}. SAVAM contains 41 fragments of 1920 $\times$ 1080 motion video from various feature movies, commercial clips, and stereo video databases. Coutrot Database 1 contains 60 videos of people or faces, moving objects, and landscapes at 720 $\times$ 576 resolution. We use the set of videos with original soundtracks for Coutrot Database 1. Ground truth saliency data collected with commercial eye trackers are provided for each video. From these videos, we manually selected 33 that we found relatively more engaging. The videos we used are \{2, 5, 6, 11, 12, 22, 24, 29, 30, 34, 38\} from SAVAM and \{6, 8, 15, 17, 18, 21, 23, 24, 25, 26, 28, 46, 47, 48, 50, 51, 52, 53, 54, 56, 57, 59\} from Coutrot Database 1. Participants view each video only once across the 10 sessions.

\subsection{Data Preprocessing}
\begin{figure}[h]
\centering

\begin{subfigure}{0.2\textwidth}
\includegraphics[width=1\linewidth]{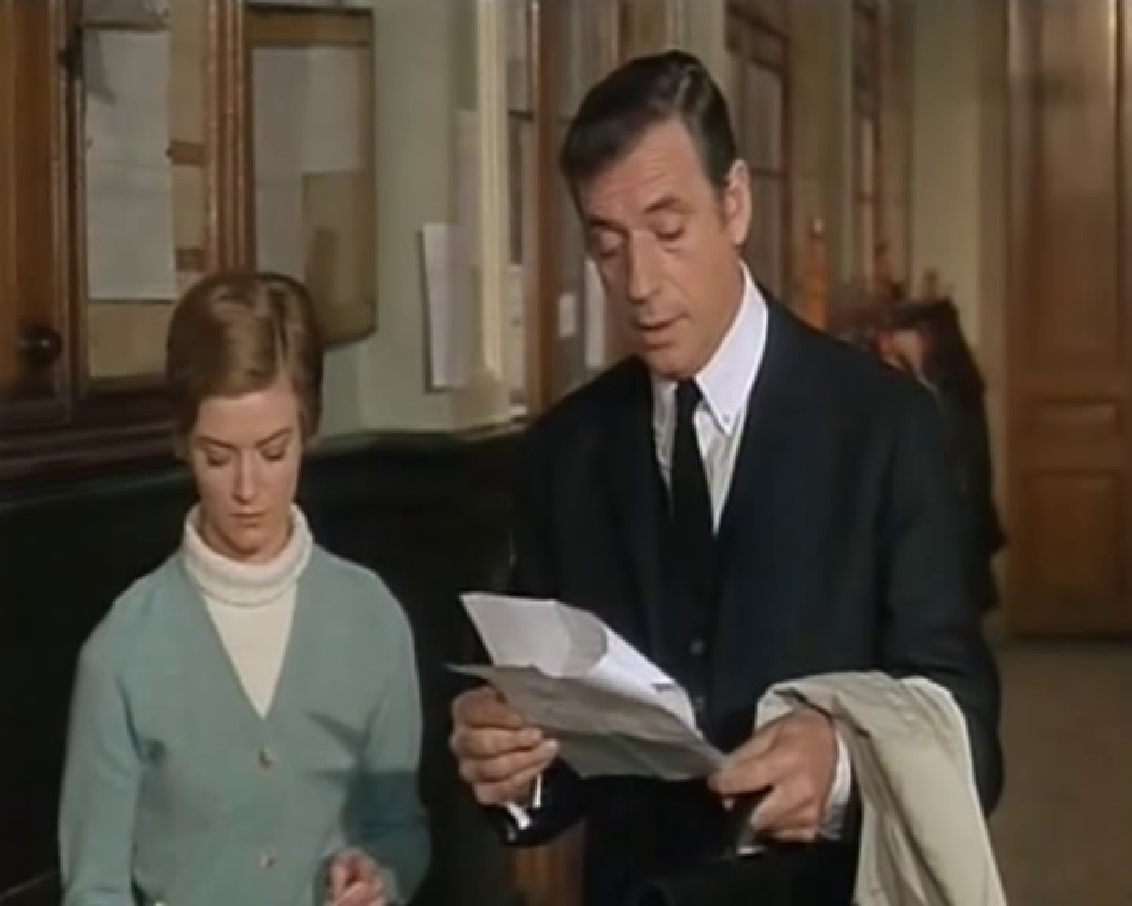} 
\caption{}
\end{subfigure}
\begin{subfigure}{0.2\textwidth}
\includegraphics[width=1\linewidth]{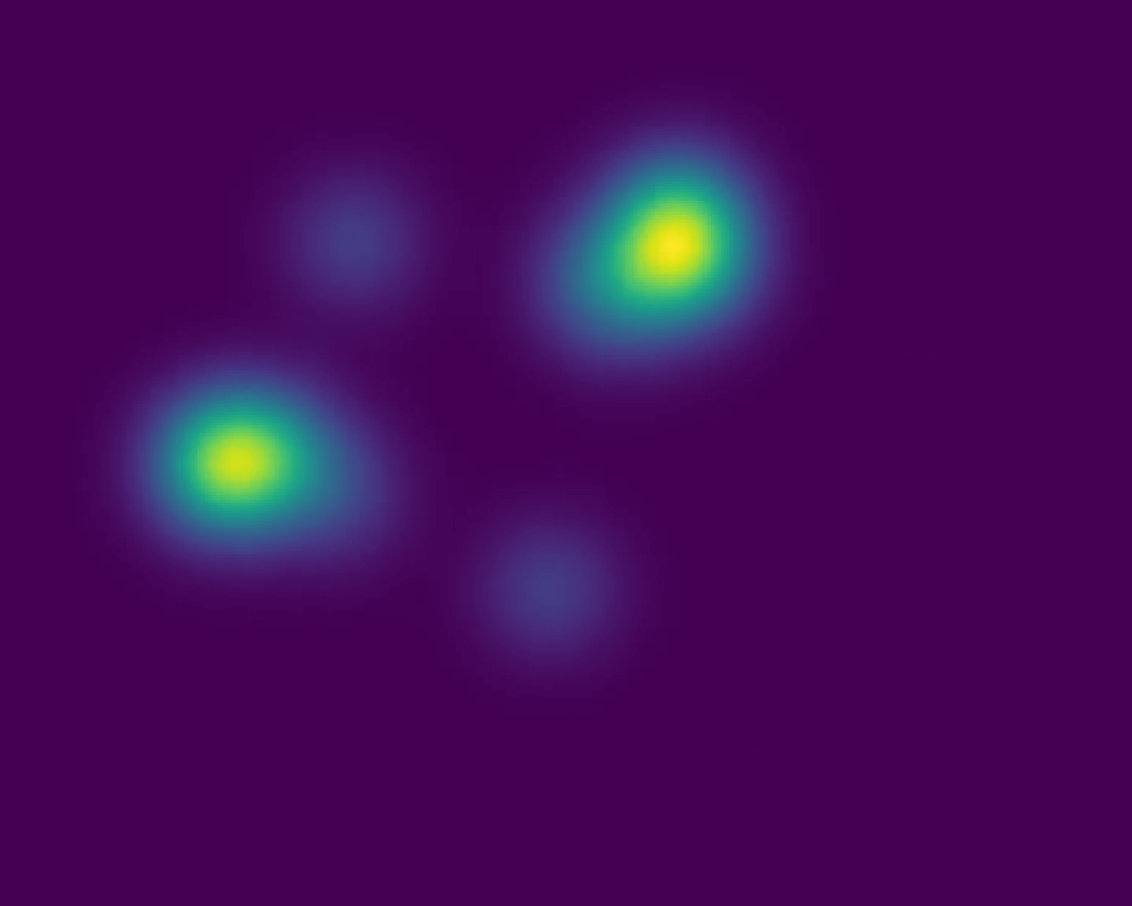}
\caption{}
\end{subfigure}

\begin{subfigure}{0.15\textwidth}
\includegraphics[width=1\linewidth]{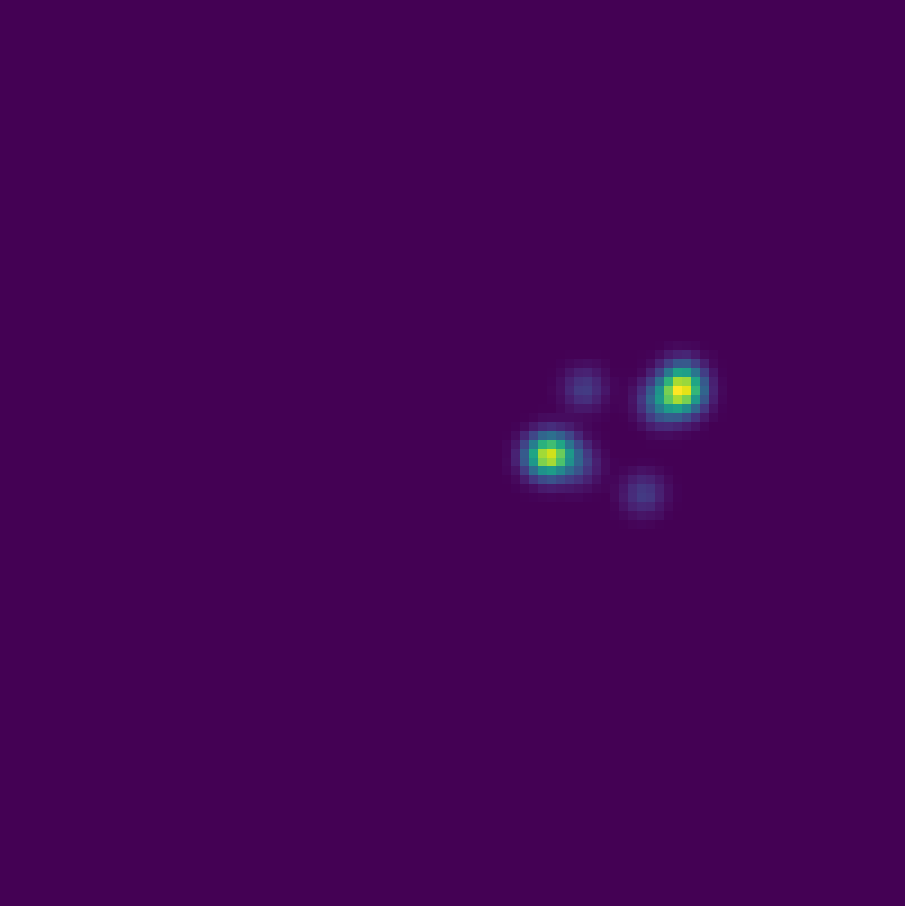}
\caption{}
\end{subfigure}
\begin{subfigure}{0.15\textwidth}
\includegraphics[width=1\linewidth]{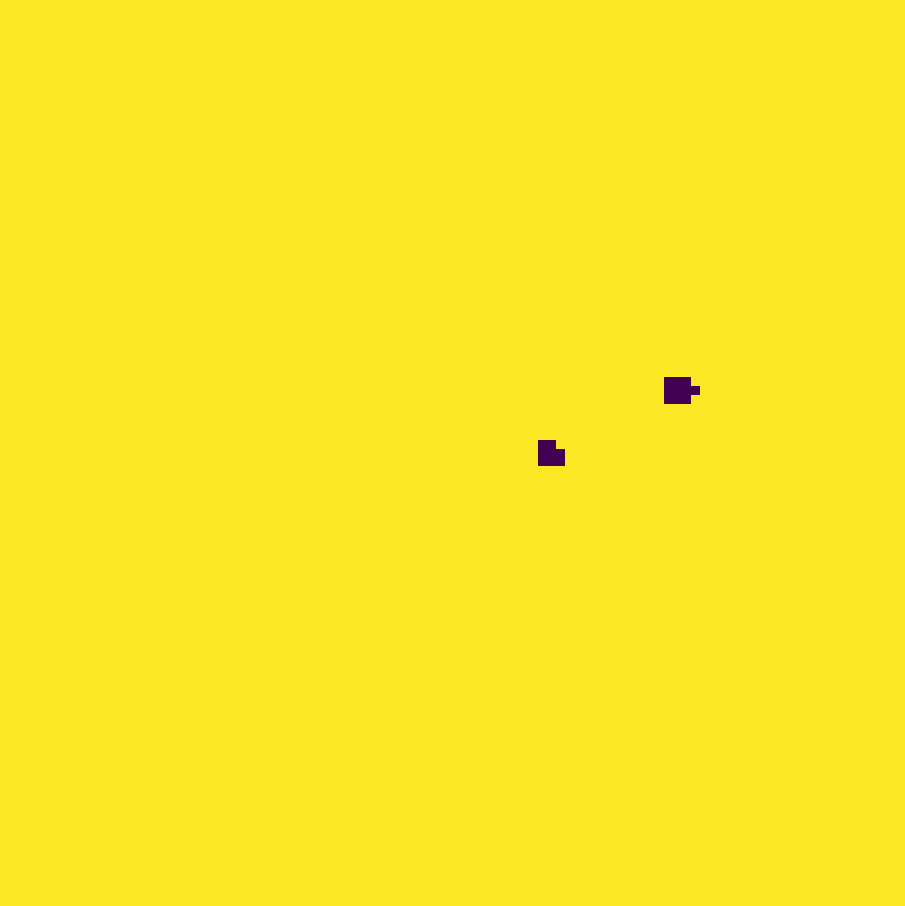}
\caption{}
\end{subfigure}
 \begin{subfigure}{0.15\textwidth}
\includegraphics[width=1\linewidth]{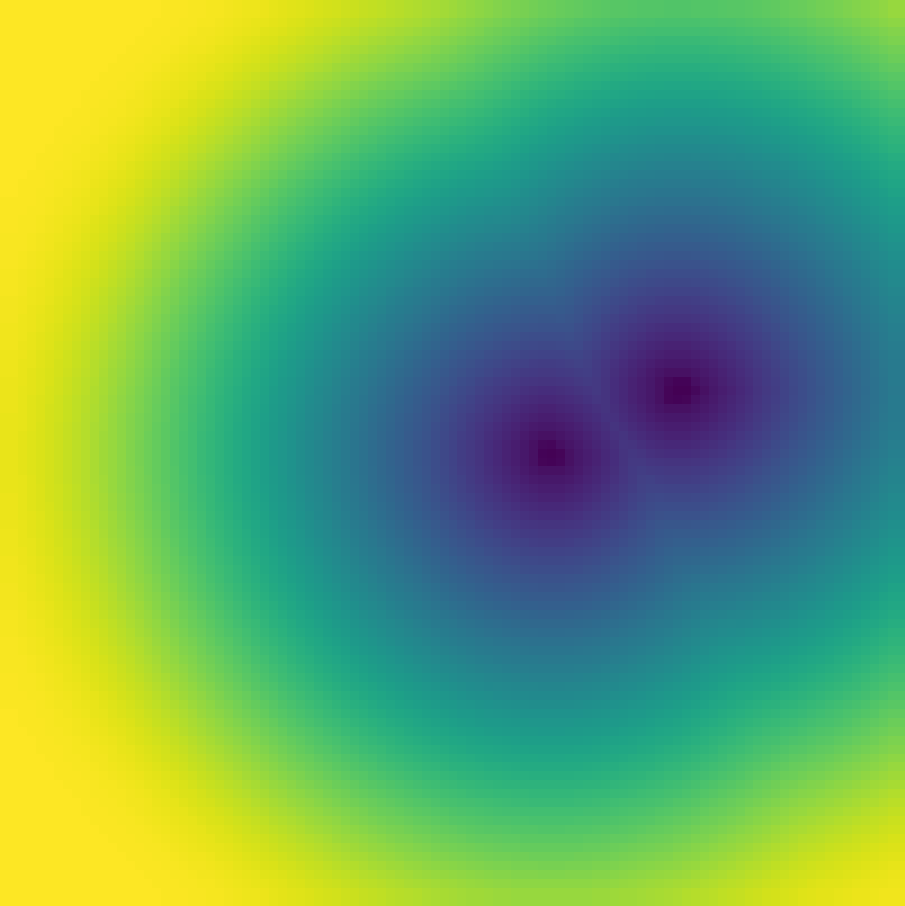}
\caption{}
\end{subfigure}

\caption{Loss map generation process. The saliency image (b) of a particular video frame (a) is mapped to a 101 $\times$ 101 grid representing the camera coordinate space (the prediction space) from -25cm to 25cm. The mapped saliency image (c) is then converted into binary image (d) after which the reinitialization algorithm is used to generate the differentiable loss map (e).}
\label{fig:preprocessing}
\end{figure}

We compute the loss map for each video frame by first mapping the saliency maps from the pixel coordinate space to the camera coordinate space defined in \cite{krafka2016eye} based on the device specifications and device orientation. We use a 101 $\times$ 101 grid to represent the square area between -25cm and 25cm in the camera coordinate space. The loss map is then computed using the method described
Section \ref{subsection:map_loss}. We set $\lambda$ equal to the 95th percentile of the saliency map for generating the binary map. This process is shown in Figure \ref{fig:preprocessing}.

The recorded videos are synced to the presented videos such that each recorded frame in the point phase is associated with the presented point location and each recorded frame in the video phase is associated with the loss map of the corresponding presented video frame. Due to high similarity between consecutive frames, we sample the recorded video and loss maps at 1 frame per second. We also discard any frame within the first 0.5 seconds of a point appearing to allow for participants to focus their gaze. Dlib \cite{dlib09} is then used to detect the face and facial landmarks in each frame. We use the face region given by Dlib to crop out the face image and compute the face grid defined in \cite{krafka2016eye}. We define the eye region as the square area centered at the midpoint of the eye corner landmarks with a side length equal to 1.8 times the horizontal distance between the eye corner landmarks. We found this results in similar eye images to \cite{krafka2016eye} which uses a native iOS algorithm to generate the croppings.

\subsection{Baseline Model} \label{section:baseline}
Our baseline model is trained using the same training and validation data from GazeCapture \cite{krafka2016eye}. We resize the input images to 64 $\times$ 64 and train for 75,000 iterations with a batch size of 256. A comparison of our baseline model to iTracker \cite{krafka2016eye} on 3 subsets of the GazeCapture test data is shown in Table \ref{table:baseline}. Due to decreasing the resolution of the input images, which we did for computational purposes, our baseline model performs slightly worse than iTracker. However, our goal here is to not to outperform iTracker but rather to set a baseline for indirect comparison with our personalized models in the following sections. While they do not report the performance on the set of test data collected on the iPad Air 2 in landscape orientation, if we extrapolate based on the trend of our baseline model, the error is approximately 3.6cm. We show, in the next section, that our low-resolution personalized model can achieve an error of 3.3cm.

\begin{table}[ht!]
\centering
\begin{tabular}{c|c|c|c} 
\hline
\multirow{3}{*}{Model} & \multicolumn{3}{|c}{Test Data Partition} \\
\cline{2-4}
& \multirow{2}{*}{All Phones} & \multirow{2}{*}{All Tablets} & iPad Air 2 \\ [0.5ex] 
& & & (Landscape) \\ 
\hline\hline
iTracker \cite{krafka2016eye} & 2.04 & 3.32 & N/A \\ 
\hline
Baseline & 2.26 & 3.76 & 4.07 \\
\hline
\end{tabular}
\caption{Mean error (cm) of our baseline model and iTracker on different test sets of GazeCapture.}
\label{table:baseline}
\end{table}

\subsection{Calibration with Saliency} 
\label{section:calibration}

We validate the usefulness of saliency data for personalized gaze estimation by fine-tuning the last two fully connected layers of our baseline model described in Section \ref{section:baseline}. For each participant, we use data from the video phase and their corresponding loss maps for fine-tuning and test on data from the point phase. We compare the effect of using the map loss with the squared map loss described in Section \ref{subsection:map_loss}. As the user is not constrained or guided to look at specific locations during the video phase, we expect the collected data to present outliers where the input image does not coincide with the saliency map. Therefore, we also explore the effect of an iterative outlier removal (IOR) technique where we remove the top 5\% of data with the largest loss from training every 2 epochs. Appropriate filtering of the input videos and saliency data is important for the proposed framework, and this simple approach has been found sufficient; we discuss more on this in the Section \ref{section:discussion}. For each participant, the baseline model is fine-tuned for 10 epochs per configuration. 

\begin{figure}[ht]
\centering
\includegraphics[width=1\linewidth]{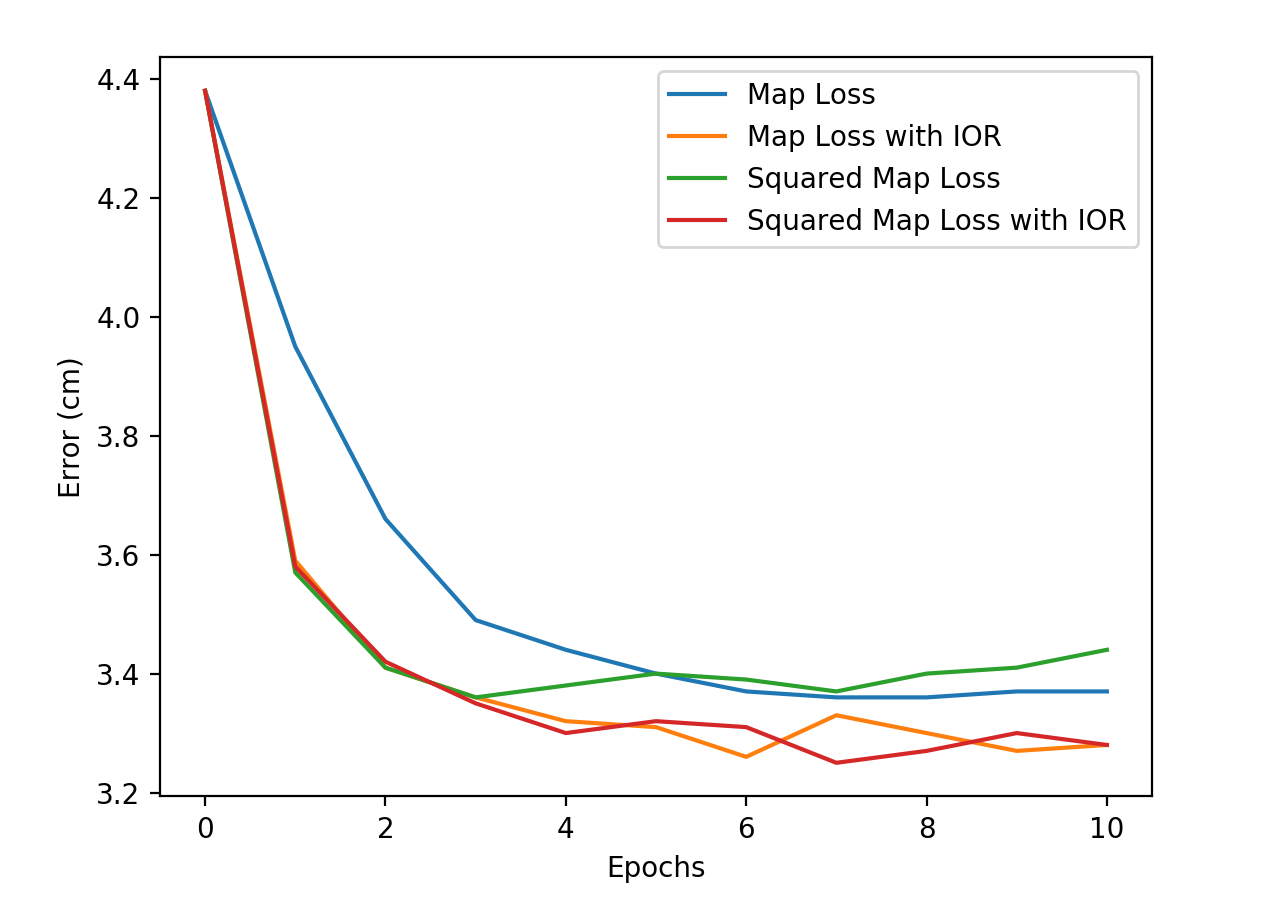} 
\caption{Average gaze error using different training configurations.}
\label{fig:calibration}
\end{figure}

Figure \ref{fig:calibration} shows the average gaze error across all participants for each epoch. It can be seen that the error stabilizes after 4 epochs. Therefore, we decide to fix the fine-tuning procedure to 4 epochs for this and subsequent experiments. Quantitative results are shown in Table \ref{table:calibration}. It can be seen that all 4 configurations improve over the baseline and IOR helps with reducing the error. As expected and commented in Section \ref{subsection:map_loss}, the squared loss leads to a faster convergence but is more susceptible to outliers with which IOR helps to mitigate. For the experiments in sections \ref{section:quantity_test} and \ref{section:generated_saliency}, we use the map loss with IOR configuration.

\begin{table}[h!]
\centering
\begin{tabular}{l|c|c} 
\hline
Configuration & Error (cm) & Improvement (\%) \\
\hline\hline
Baseline & 4.38 $\pm$ 1.18 & 0 \\ 
\hline
Map Loss & 3.44 $\pm$ 0.74 & 21.46 \\
\hline
Map Loss w/ IOR & 3.32 $\pm$ 0.77 & 24.20 \\
\hline
Squared Map Loss & 3.38 $\pm$ 0.73 & 22.83 \\
\hline
Squared Map Loss & \multirow{2}{*}{3.30 $\pm$ 0.78} &  \multirow{2}{*}{24.66} \\
w/ IOR & & \\
\hline
\end{tabular}
\caption{Average gaze error using different training configurations.}
\label{table:calibration}
\end{table}

\subsection{Scalability with Data Quantity} 
\label{section:quantity_test}
Contrary to standard point-wise calibration, using visual saliency for calibration is transparent and therefore provide a means for collecting large amounts of data in a non-intrusive way. To examine how the performance of our approach scales with the amount of data, we use data from 1, 2, 4, and 8 of the 10 sessions to fine-tune the baseline model for each participant. Similar to Section \ref{section:baseline}, fine-tuning is performed on data from the video phase of the selected sessions, and testing is performed on data from the point phase of all 10 sessions. Quantitative results are shown in Table \ref{table:quantity_test}. It can be seen that the estimation error decreases with the increase of training data.

\begin{table}[h]
\centering
\begin{tabular}{c|c|c} 
\hline
Sessions of data & \multirow{2}{*}{Error (cm)} & \multirow{2}{*}{Improvement (\%)} \\
used for training &  & \\ [0.5ex] 
\hline\hline
Baseline & 4.38 $\pm$ 1.18 & 0 \\ 
\hline
1 & 3.73 $\pm$ 0.77 & 14.84 \\ 
\hline
2 & 3.62 $\pm$ 0.77 & 17.35 \\
\hline
4 & 3.46 $\pm$ 0.78 & 21.00 \\
\hline
8 & 3.38 $\pm$ 0.71 & 22.83 \\
\hline
\end{tabular}
\caption{Gaze estimation error using different amounts of training data.}
\label{table:quantity_test}
\end{table}

\subsection{Generated vs. Empirical Saliency}
\label{section:generated_saliency}

In the above experiments, the saliency data we use for fine-tuning were collected by \cite{coutrot2013toward, lyudvichenko2017semiautomatic} using commercial eye trackers. While this gives us reliable data, it also limits us to using videos of which ground truth saliency is available. Having the capability to use arbitrary videos would greatly broaden the applicability of our method. In this experiment, we explore the potential of using saliency prediction algorithms to generate saliency maps for gaze estimation. Specifically, we examine two state-of-the-art saliency prediction algorithms: SAM \cite{cornia2018predicting}, an image saliency prediction algorithm; and ACL \cite{wang2018revisiting}, a video saliency prediction algorithm. We use the ResNet version of SAM that is trained on the 2015 SALICON \cite{jiang2015salicon} dataset. ACL is trained on Hollywood-2 \cite{mathe2014actions}, UCF sports \cite{mathe2014actions}, and DHF1K \cite{wang2018revisiting} datasets. Both SAM and ACL models are publicly available. We independently run SAM and ACL on the video clips to generate their saliency data. We refer to the saliency data collected from eye trackers as empirical saliency data. Examples of some video frames and their respective saliency maps are shown in Figure \ref{fig:saliency_data}.

\def \myspace {0.11}

\begin{figure}[h]
\centering
\begin{subfigure}{\myspace\textwidth}
\includegraphics[width=1\linewidth]{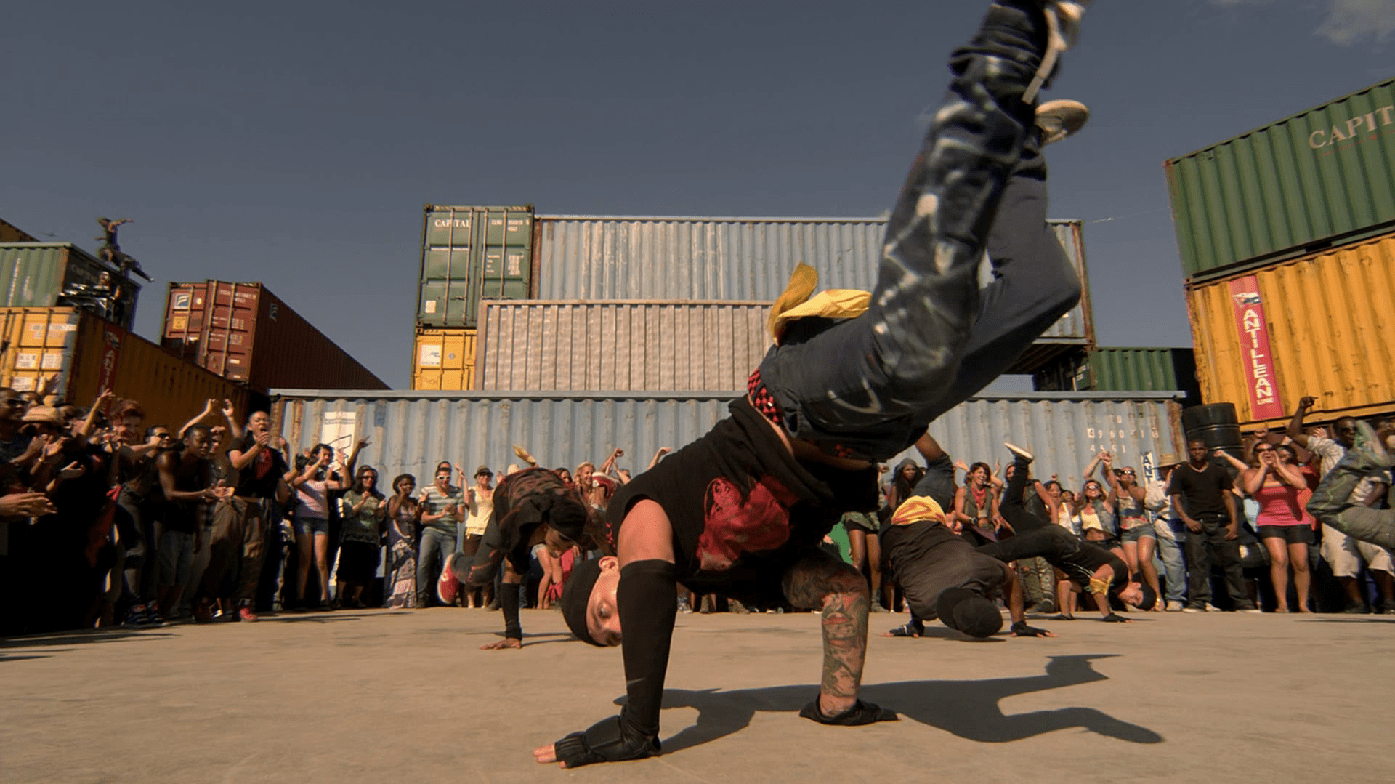} 
\end{subfigure}
\begin{subfigure}{\myspace\textwidth}
\includegraphics[width=1\linewidth]{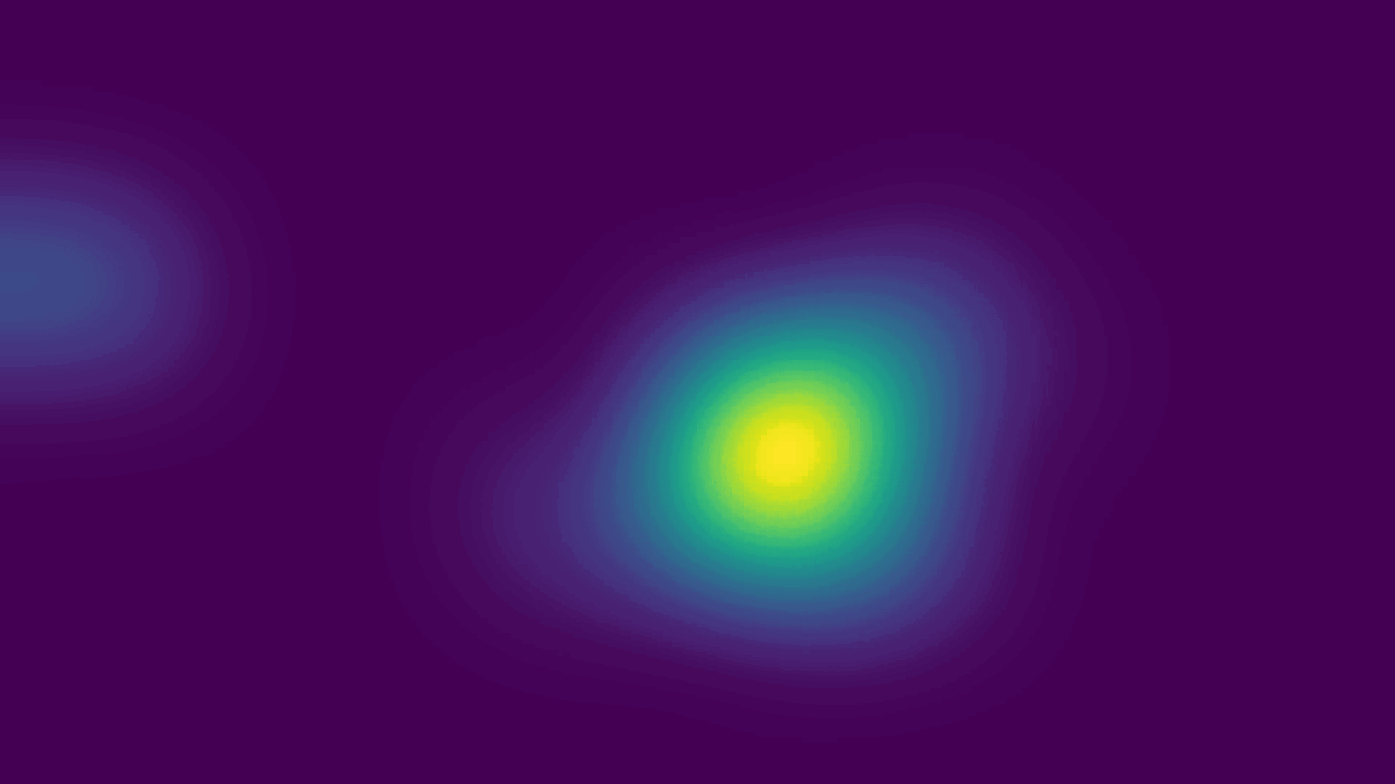}
\end{subfigure}
\begin{subfigure}{\myspace\textwidth}
\includegraphics[width=1\linewidth]{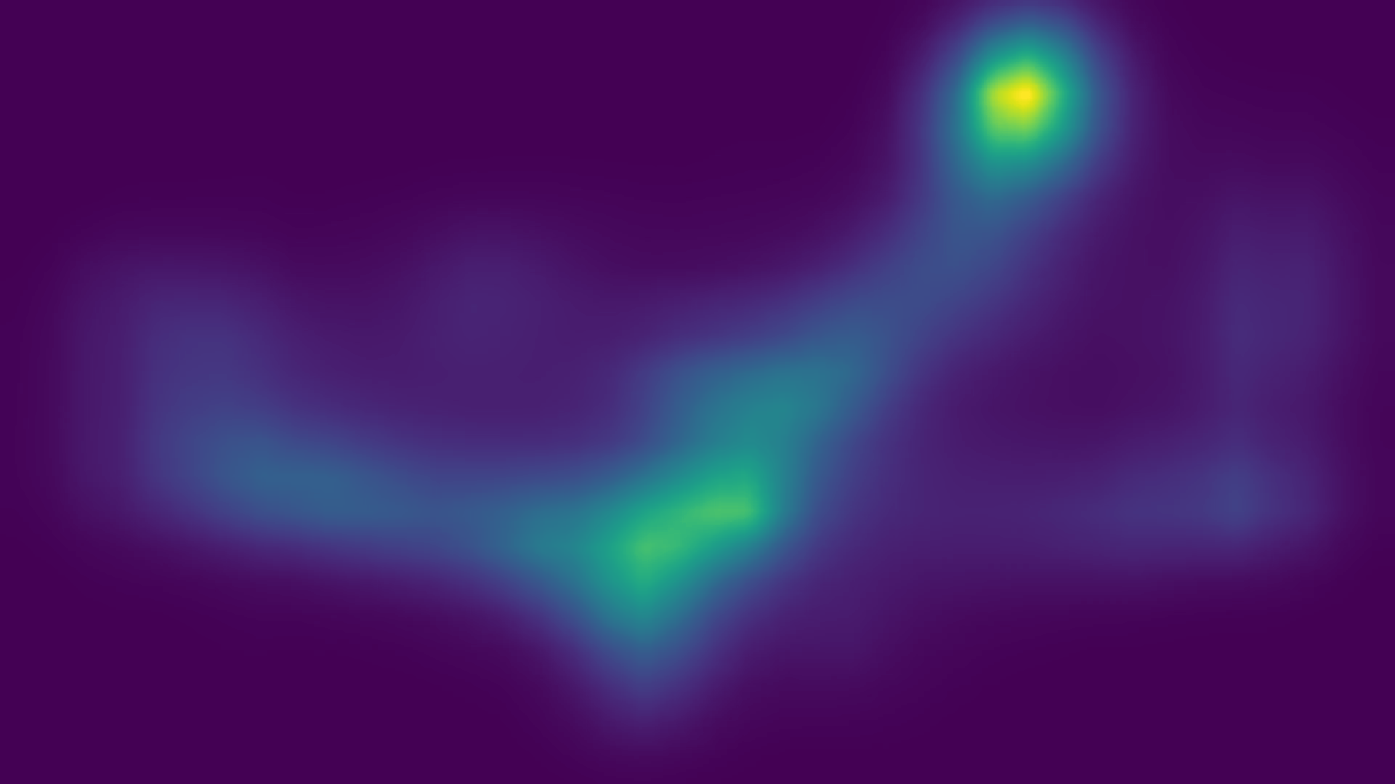}
\end{subfigure}
\begin{subfigure}{\myspace\textwidth}
\includegraphics[width=1\linewidth]{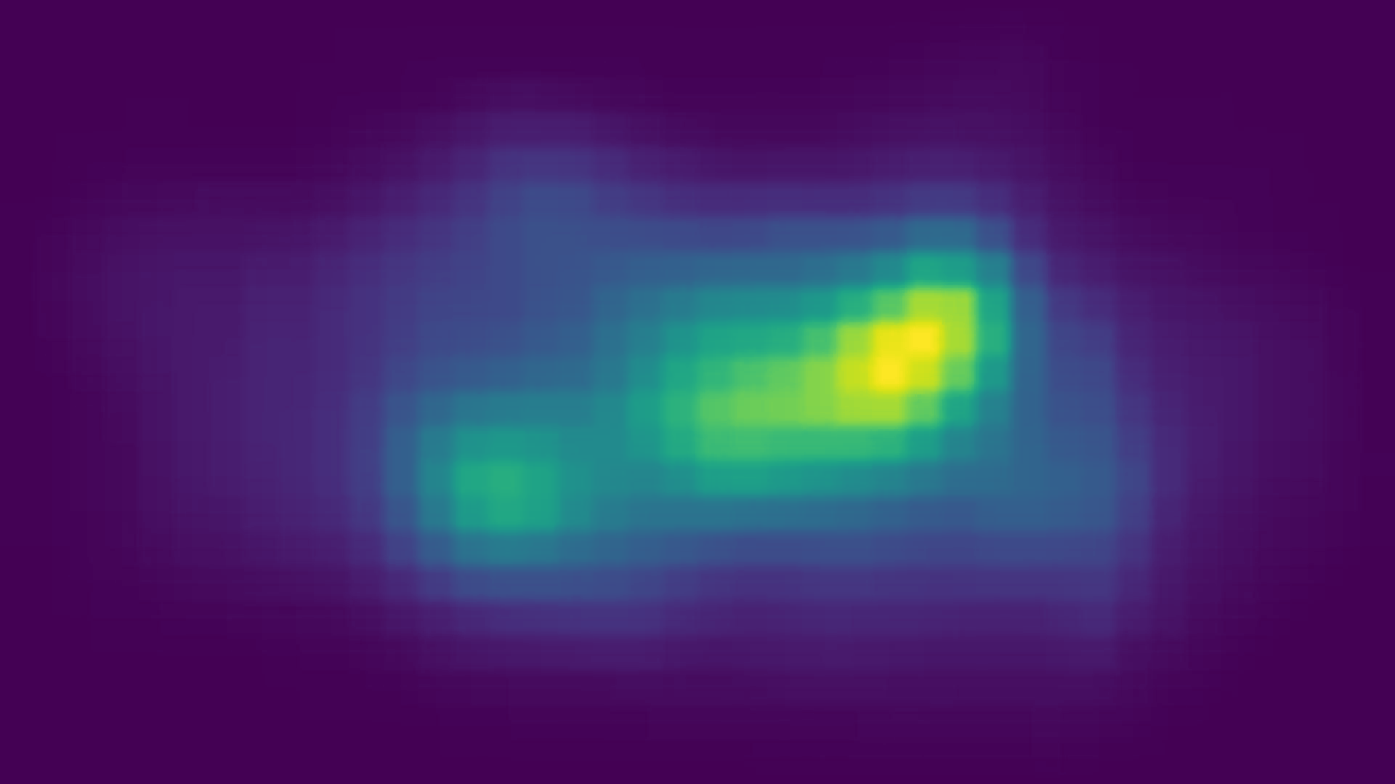}
\end{subfigure}

\begin{subfigure}{\myspace\textwidth}
\includegraphics[width=1\linewidth]{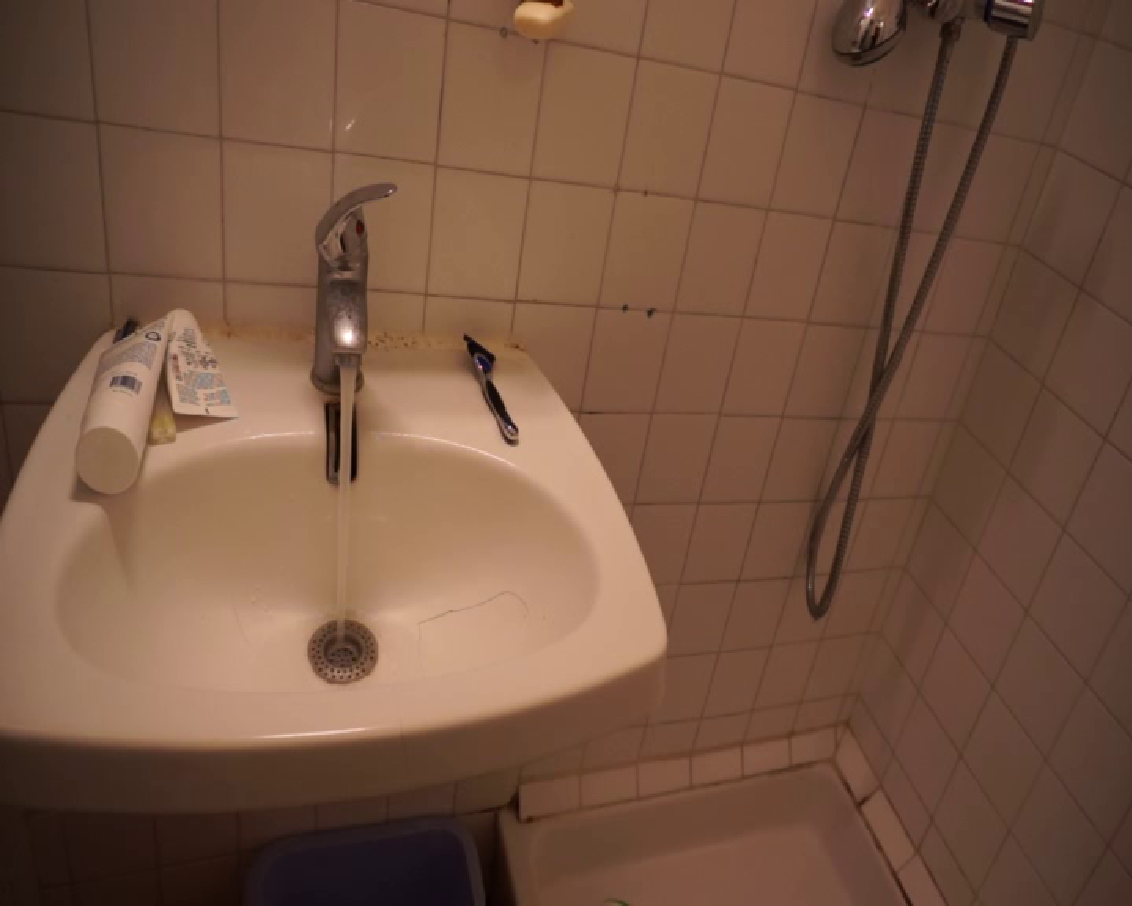} 
\end{subfigure}
\begin{subfigure}{\myspace\textwidth}
\includegraphics[width=1\linewidth]{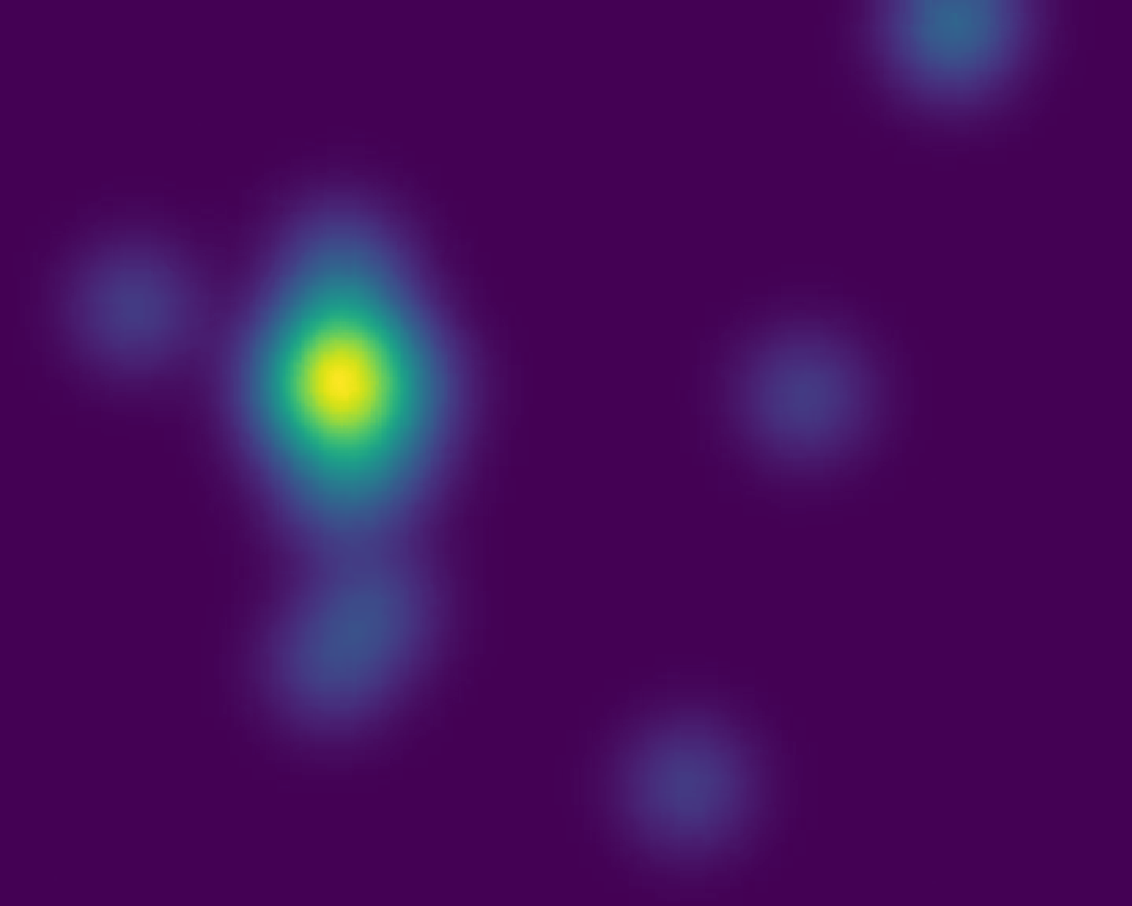}
\end{subfigure}
\begin{subfigure}{\myspace\textwidth}
\includegraphics[width=1\linewidth]{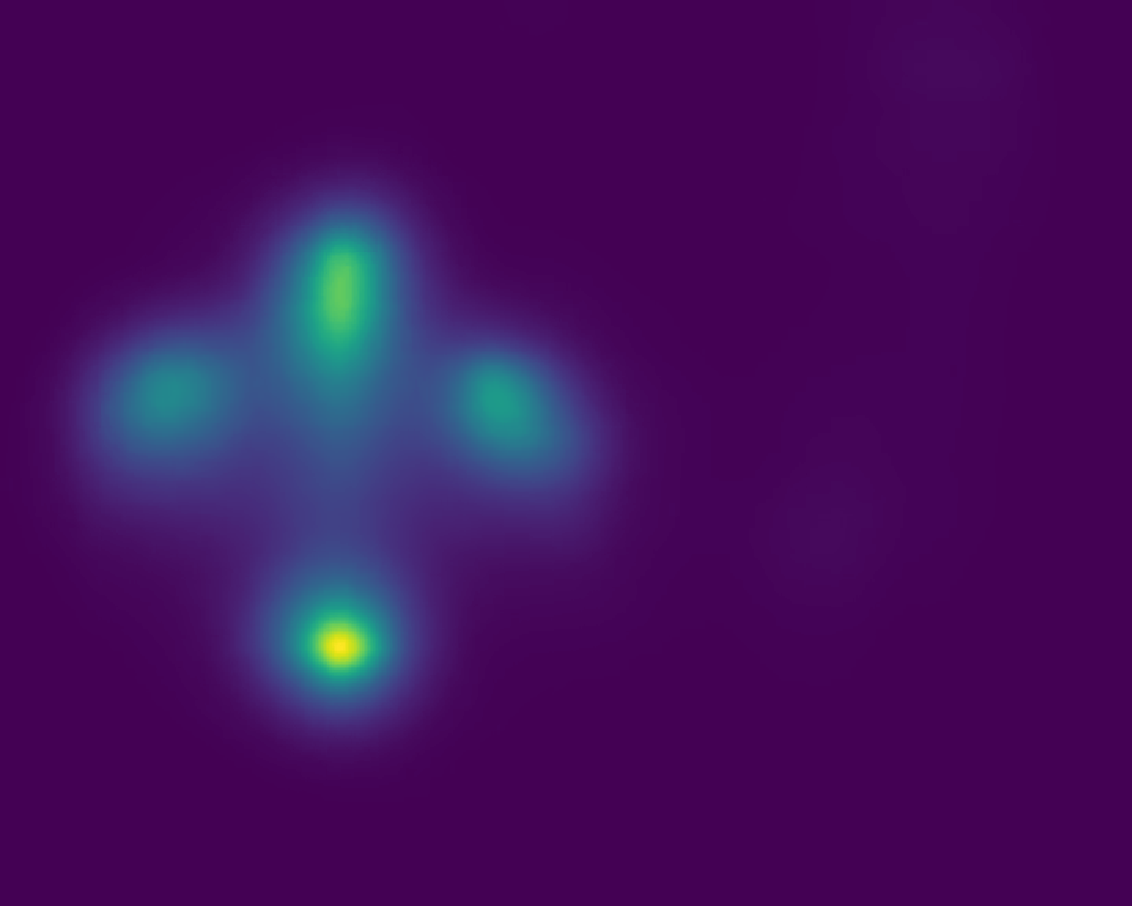}
\end{subfigure}
\begin{subfigure}{\myspace\textwidth}
\includegraphics[width=1\linewidth]{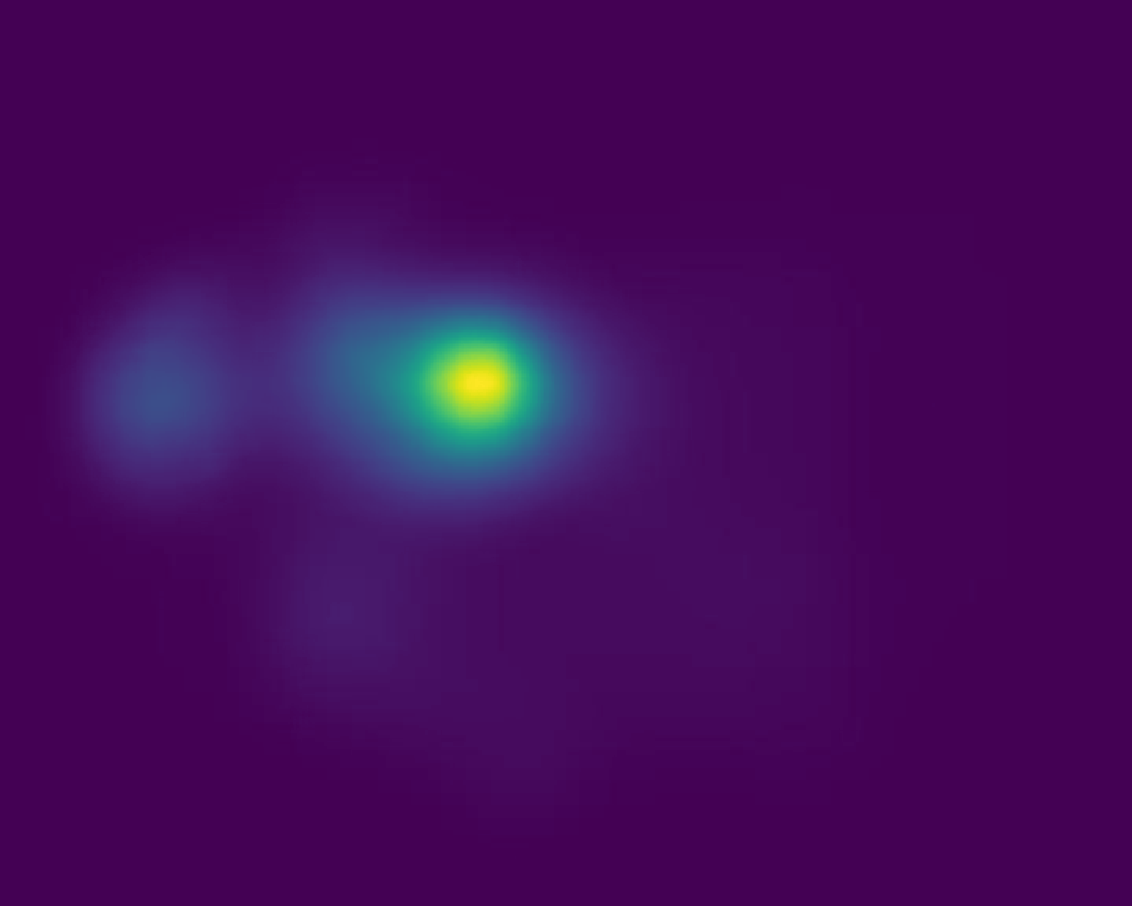}
\end{subfigure}

\begin{subfigure}{\myspace\textwidth}
\includegraphics[width=1\linewidth]{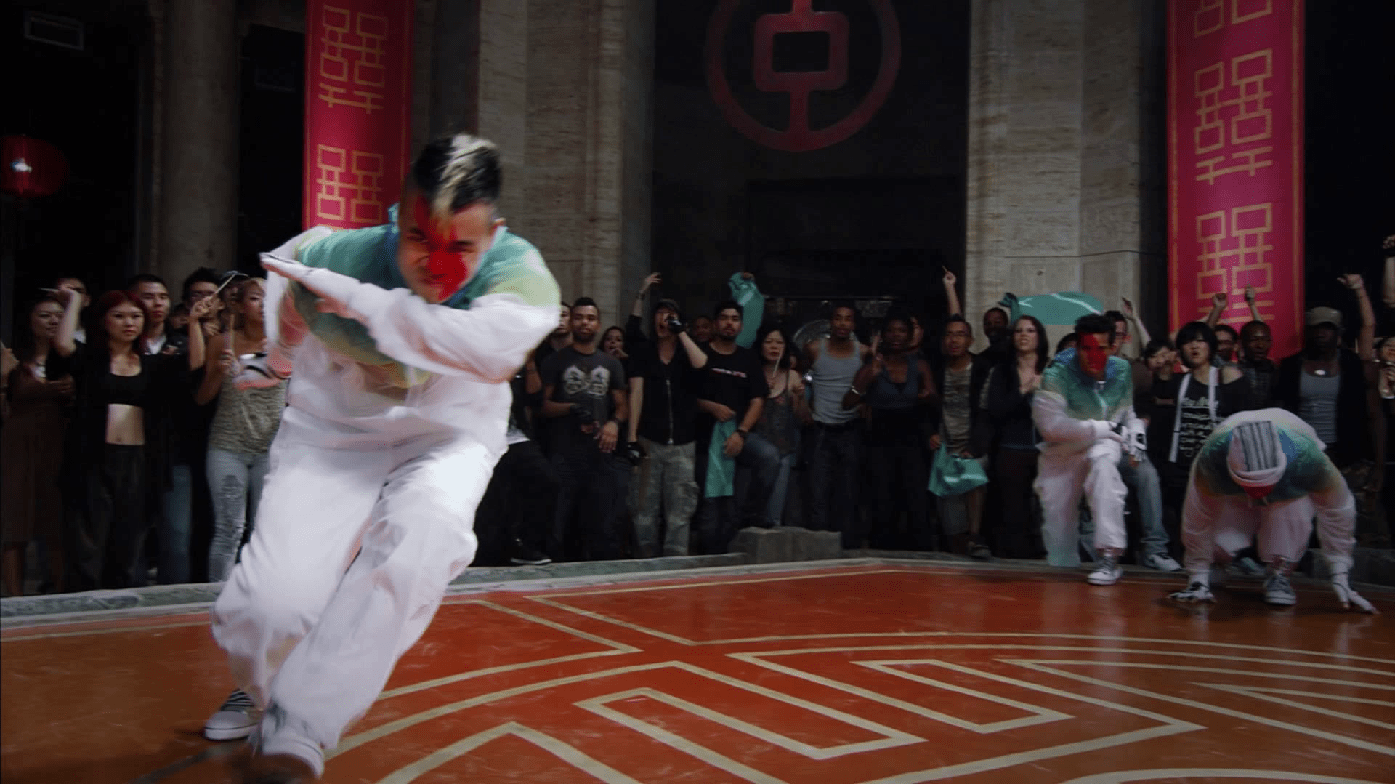} 
\end{subfigure}
\begin{subfigure}{\myspace\textwidth}
\includegraphics[width=1\linewidth]{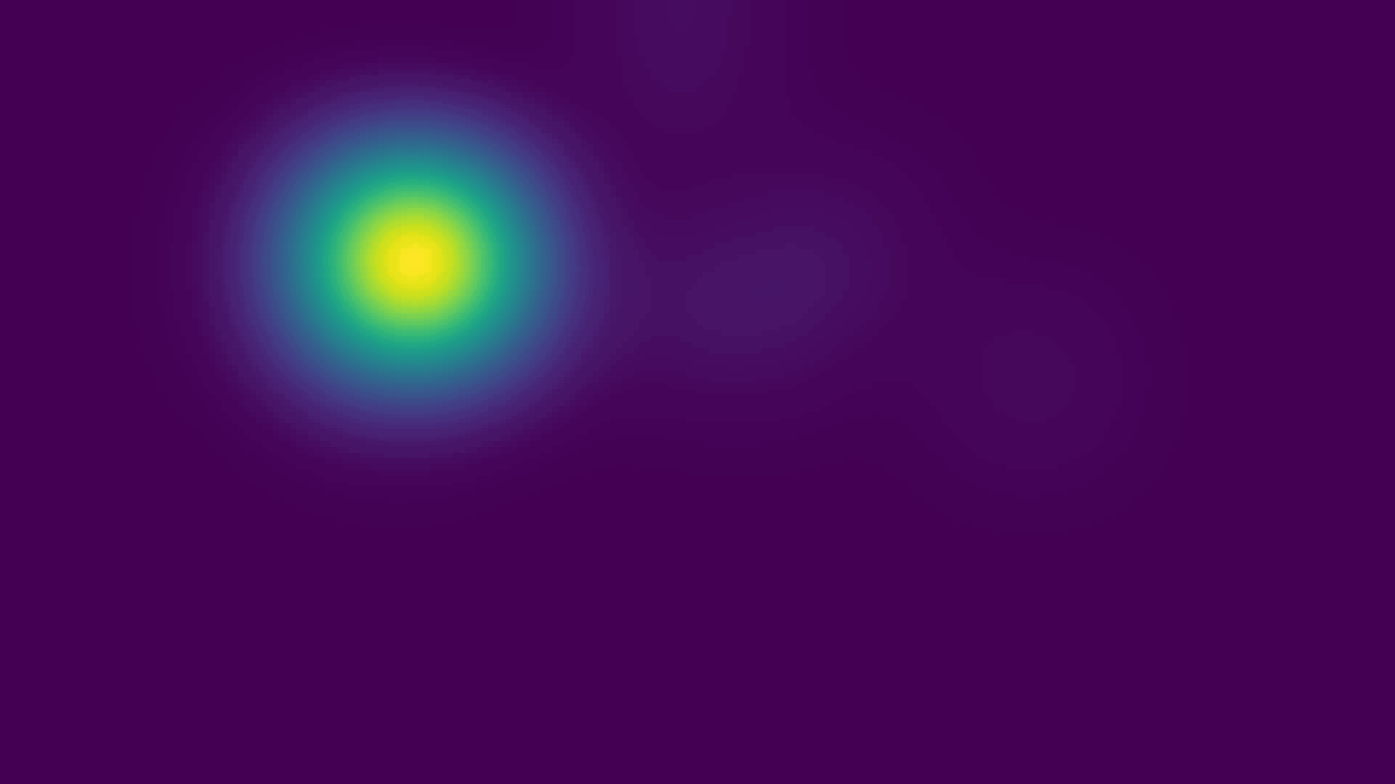}
\end{subfigure}
\begin{subfigure}{\myspace\textwidth}
\includegraphics[width=1\linewidth]{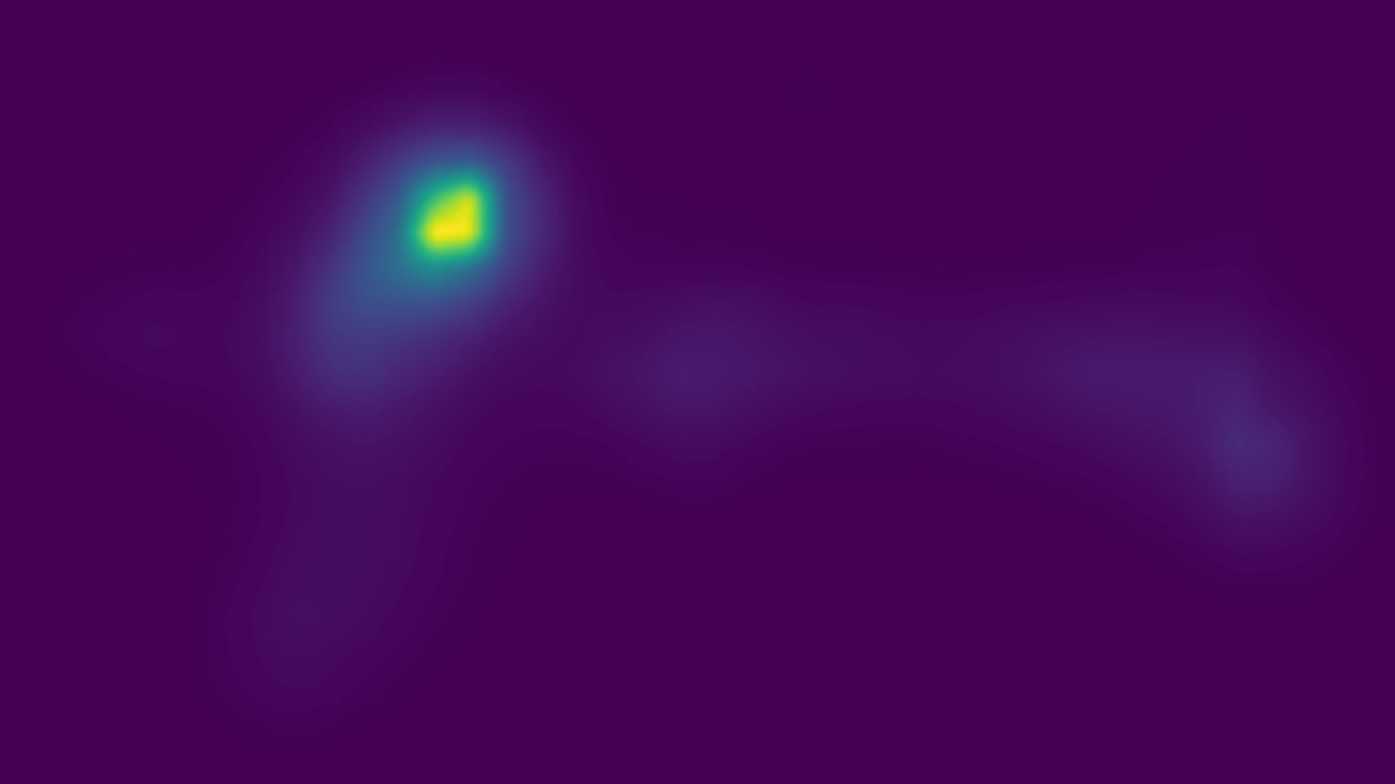}
\end{subfigure}
\begin{subfigure}{\myspace\textwidth}
\includegraphics[width=1\linewidth]{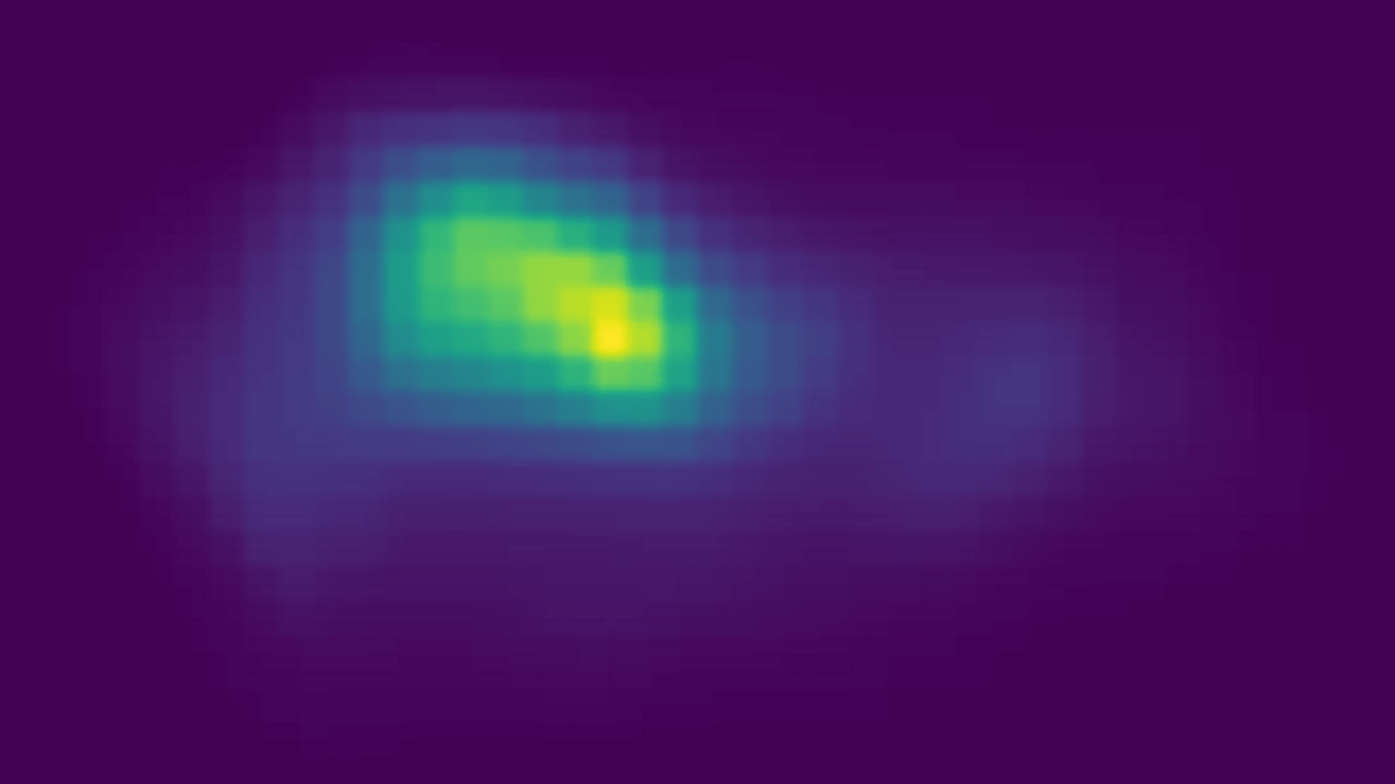}
\end{subfigure}
 
 \begin{subfigure}{\myspace\textwidth}
\includegraphics[width=1\linewidth]{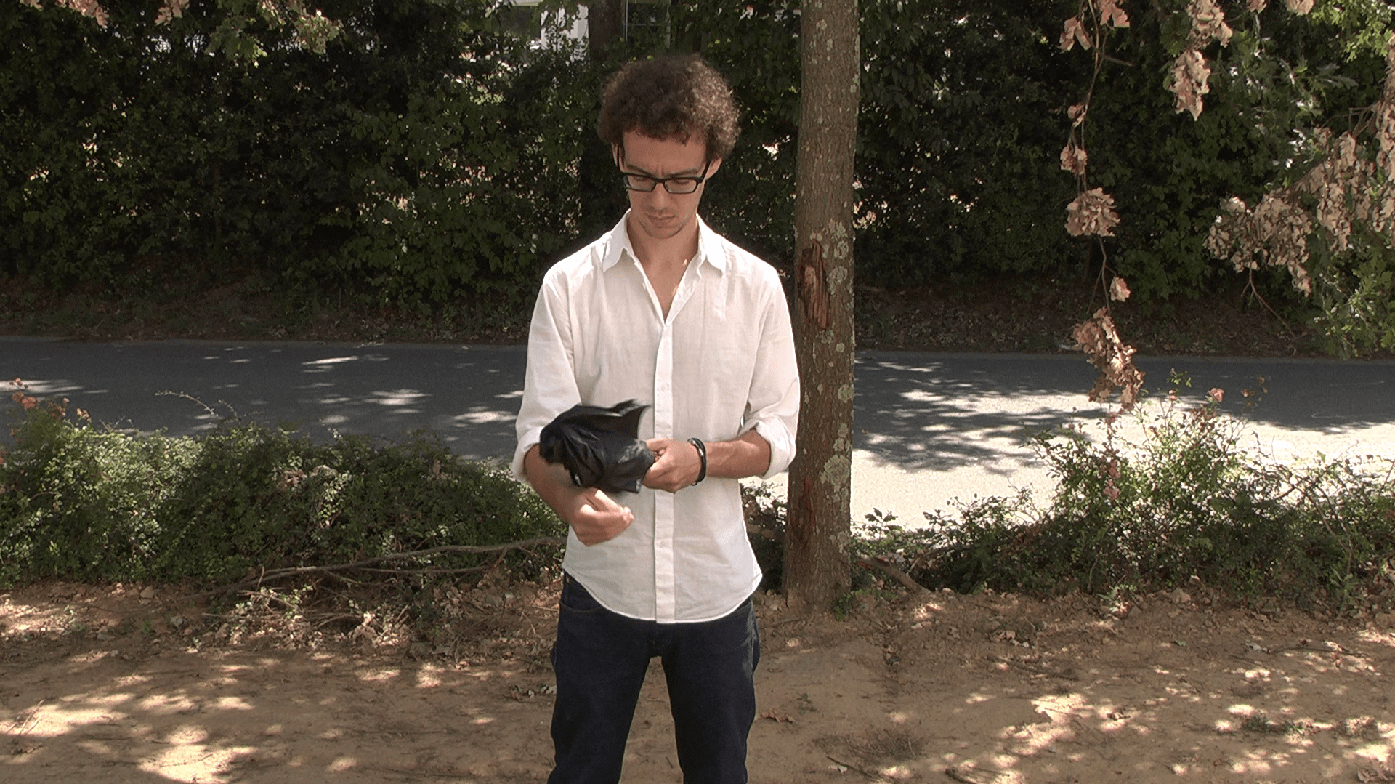} 
\end{subfigure}
\begin{subfigure}{\myspace\textwidth}
\includegraphics[width=1\linewidth]{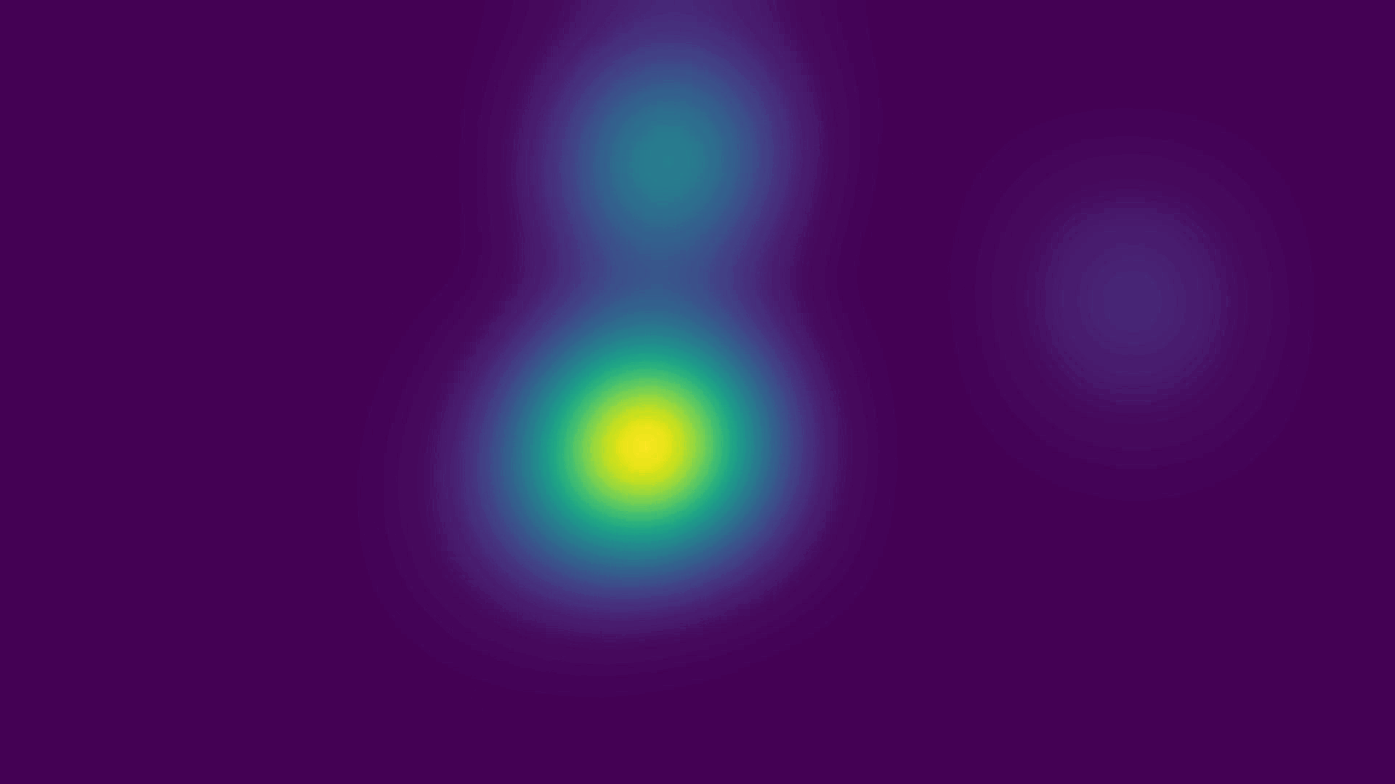}
\end{subfigure}
\begin{subfigure}{\myspace\textwidth}
\includegraphics[width=1\linewidth]{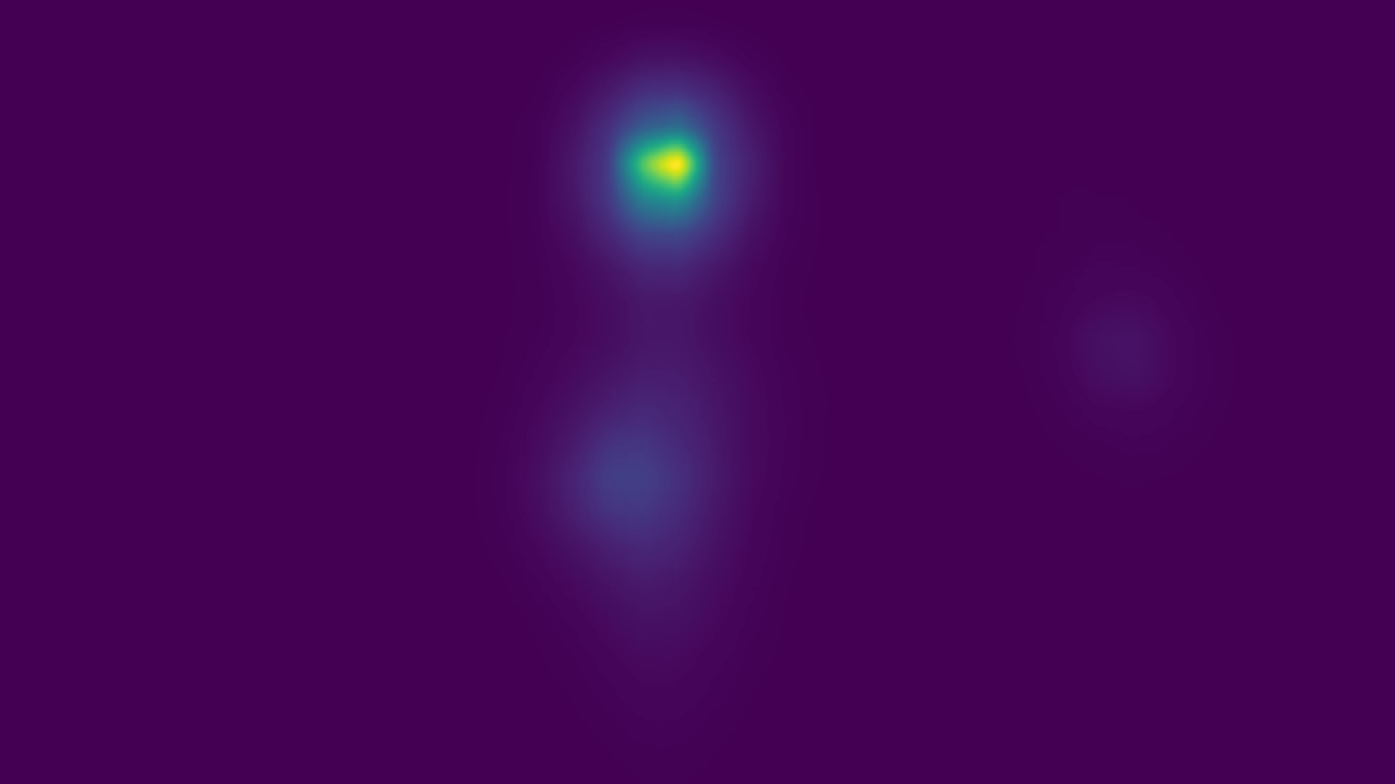}
\end{subfigure}
\begin{subfigure}{\myspace\textwidth}
\includegraphics[width=1\linewidth]{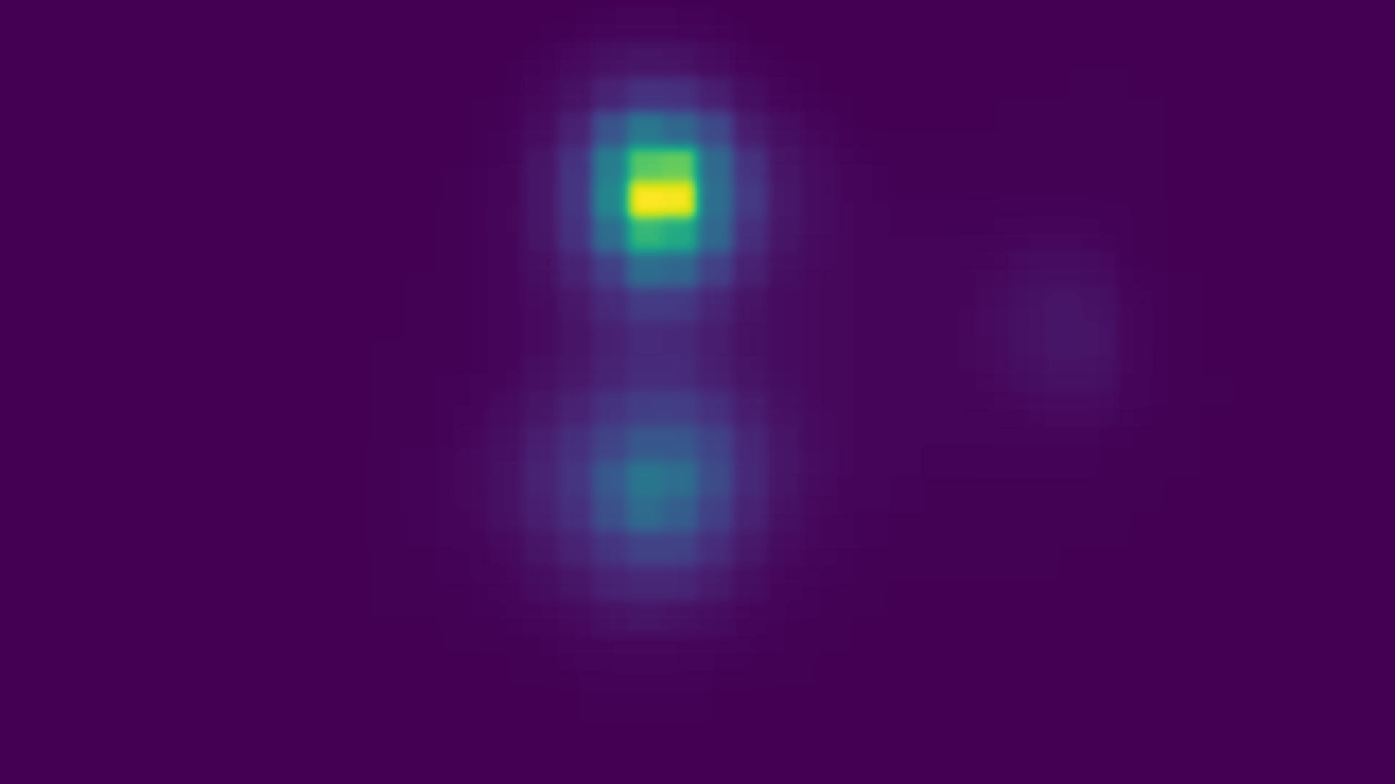}
\end{subfigure}

\begin{subfigure}{\myspace\textwidth}
\includegraphics[width=1\linewidth]{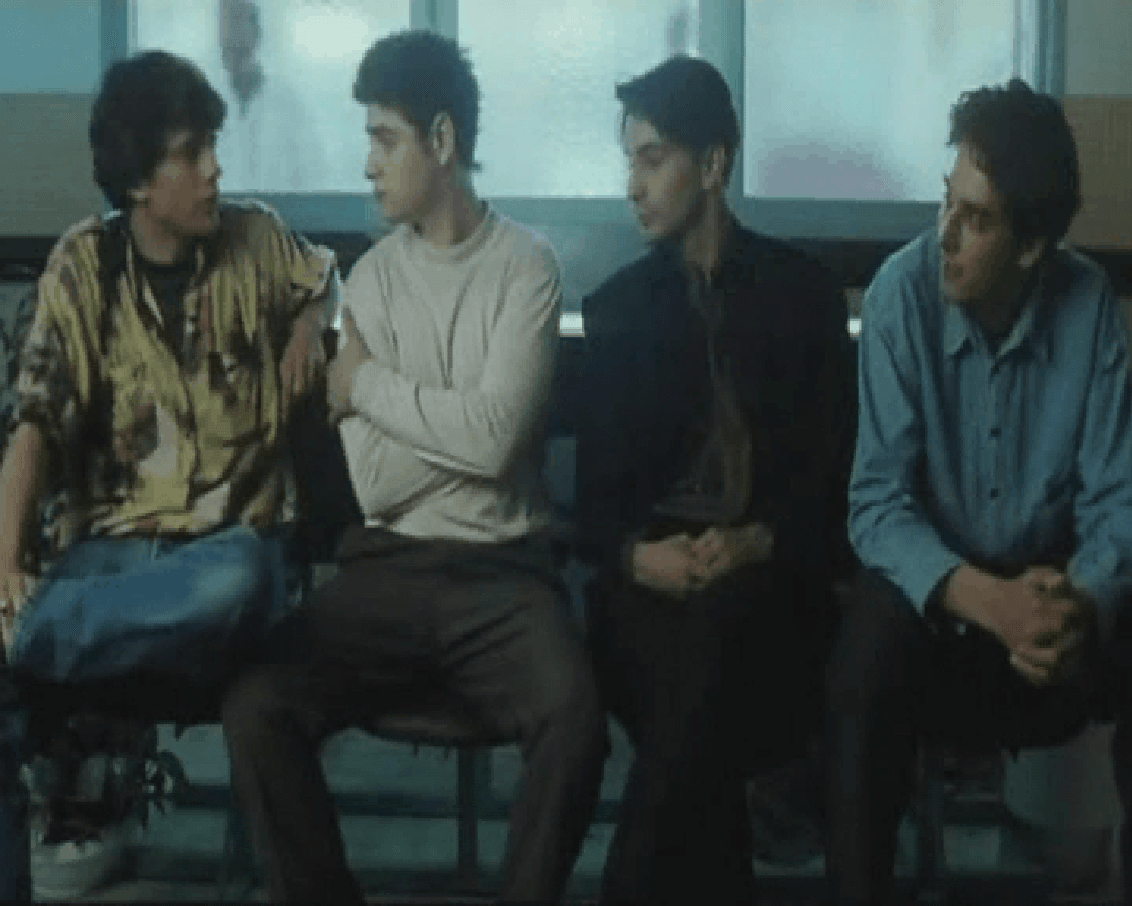} 
\end{subfigure}
\begin{subfigure}{\myspace\textwidth}
\includegraphics[width=1\linewidth]{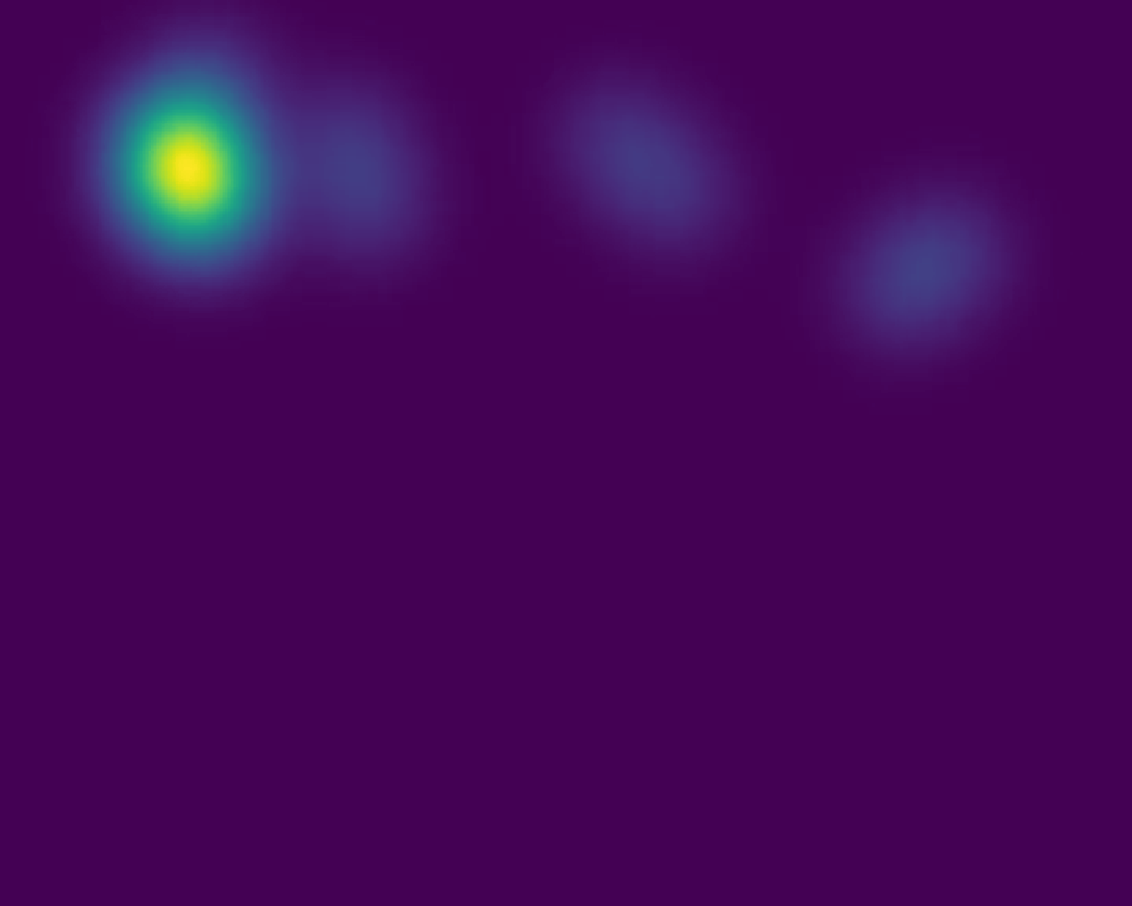}
\end{subfigure}
\begin{subfigure}{\myspace\textwidth}
\includegraphics[width=1\linewidth]{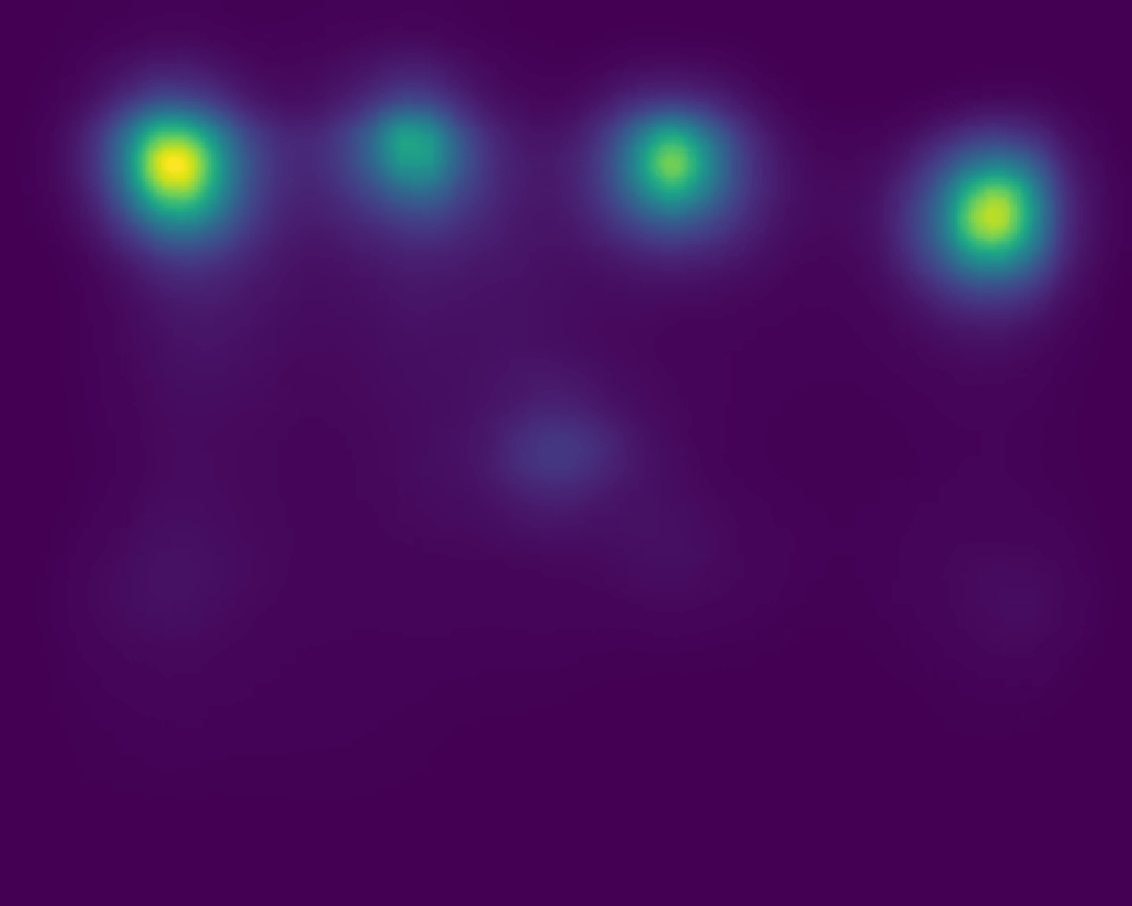}
\end{subfigure}
\begin{subfigure}{\myspace\textwidth}
\includegraphics[width=1\linewidth]{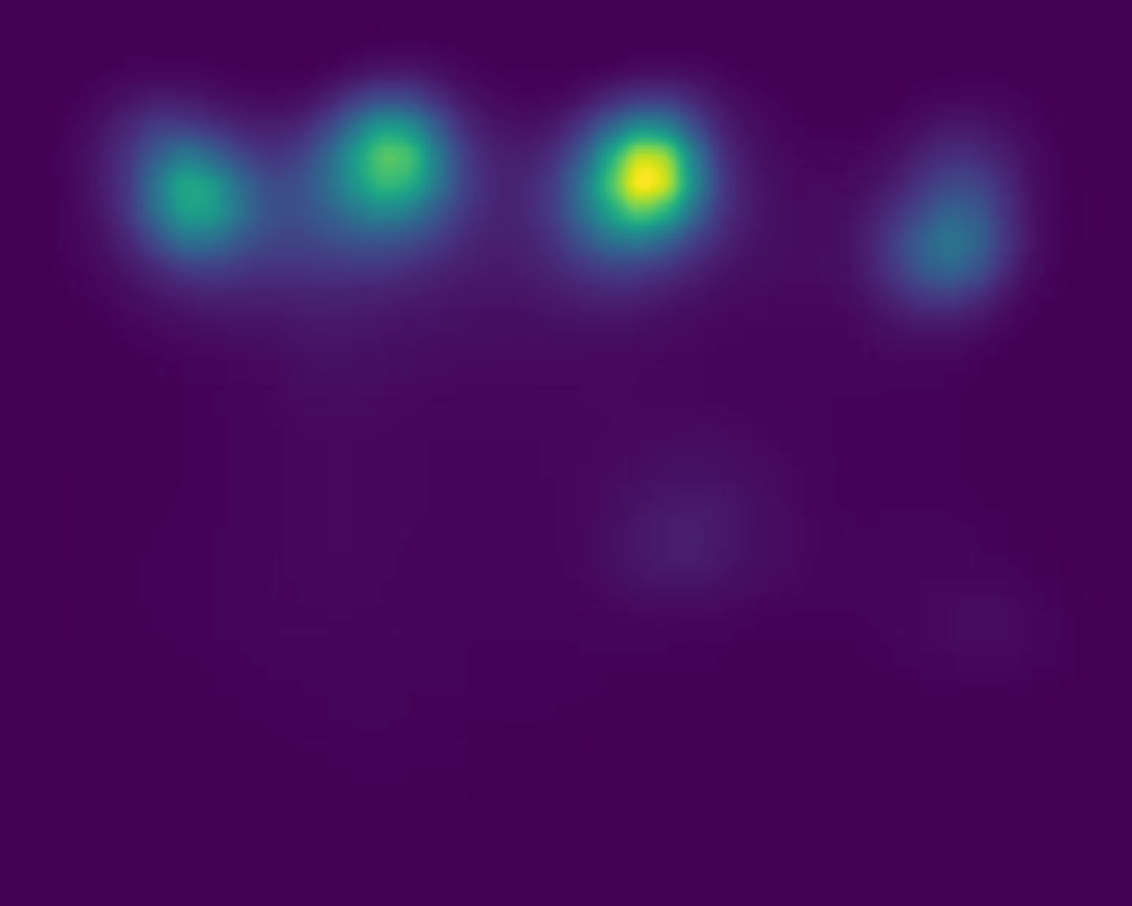}
\end{subfigure}

\begin{subfigure}{\myspace\textwidth}
\includegraphics[width=1\linewidth]{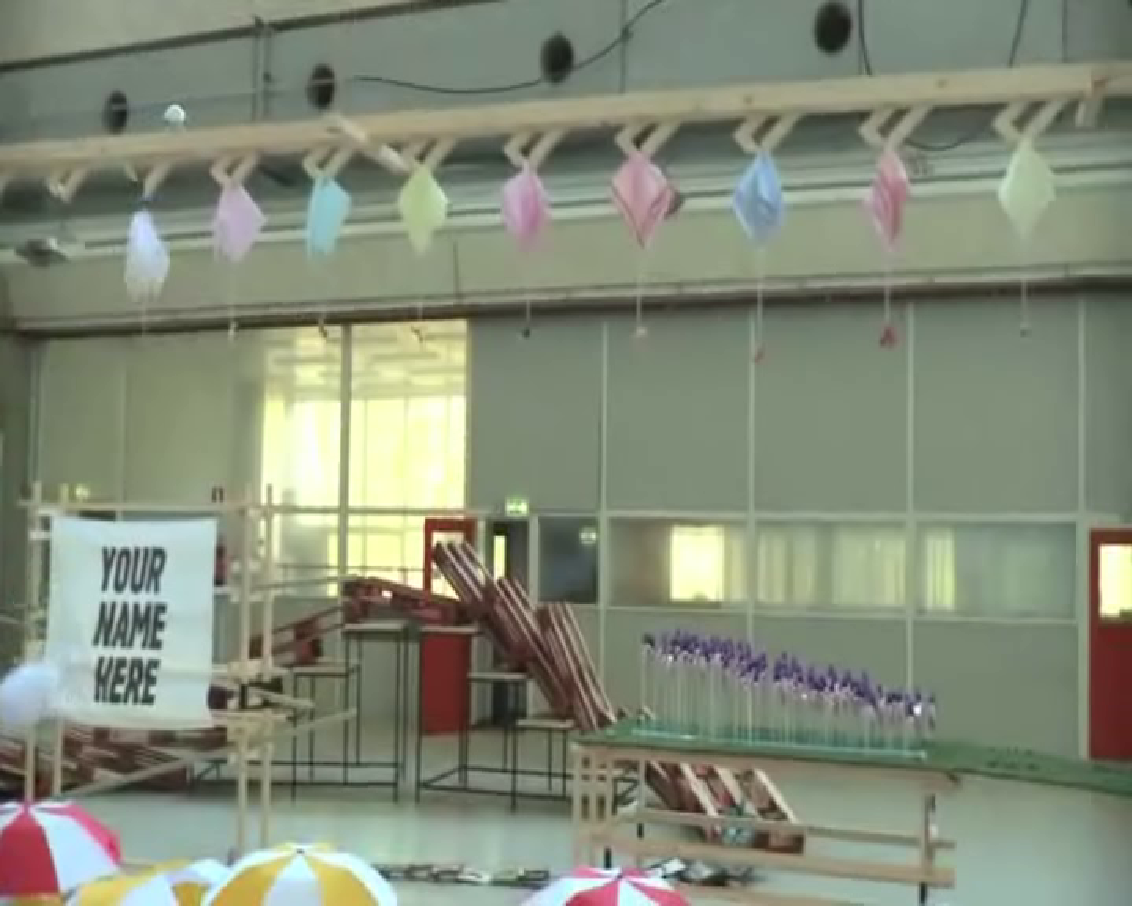} 
\captionsetup{labelformat=empty}
\caption{Video}
\end{subfigure}
\begin{subfigure}{\myspace\textwidth}
\includegraphics[width=1\linewidth]{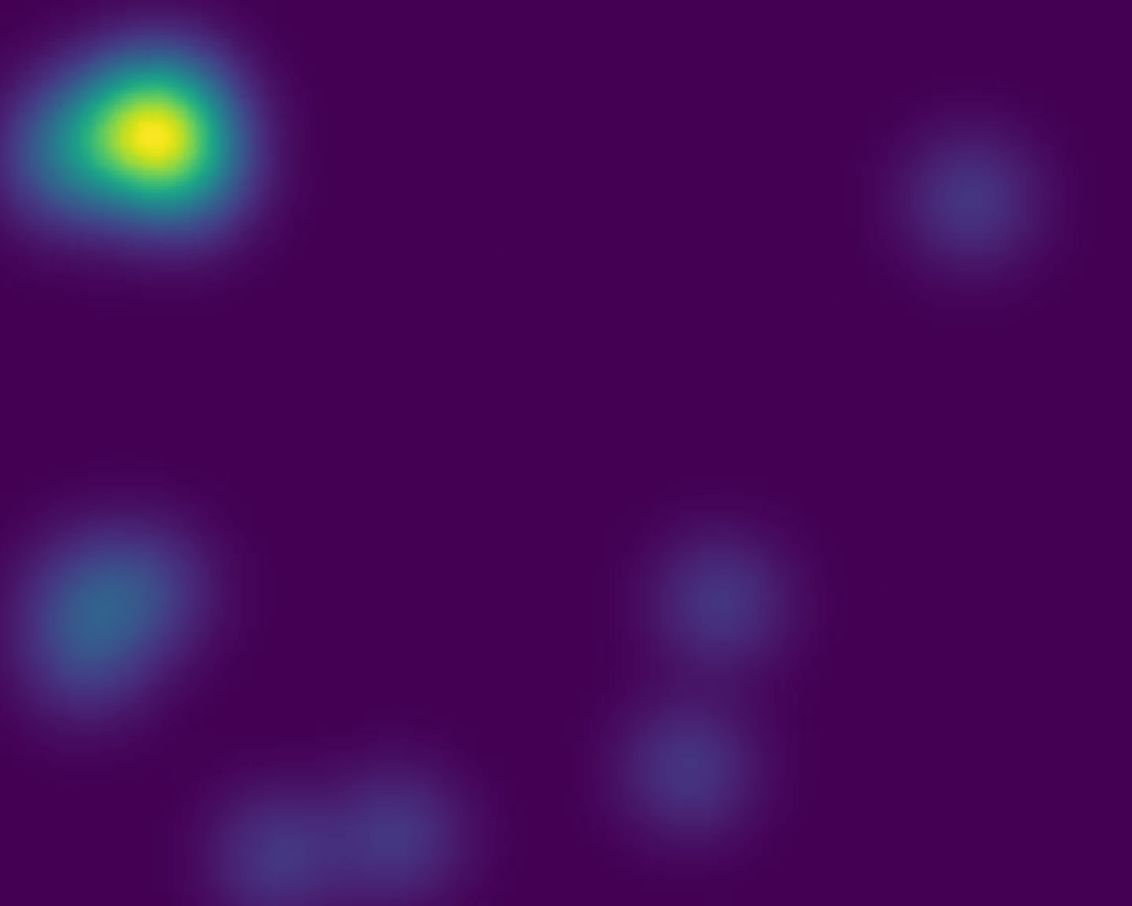}
\captionsetup{labelformat=empty}
\caption{Empirical}
\end{subfigure}
\begin{subfigure}{\myspace\textwidth}
\includegraphics[width=1\linewidth]{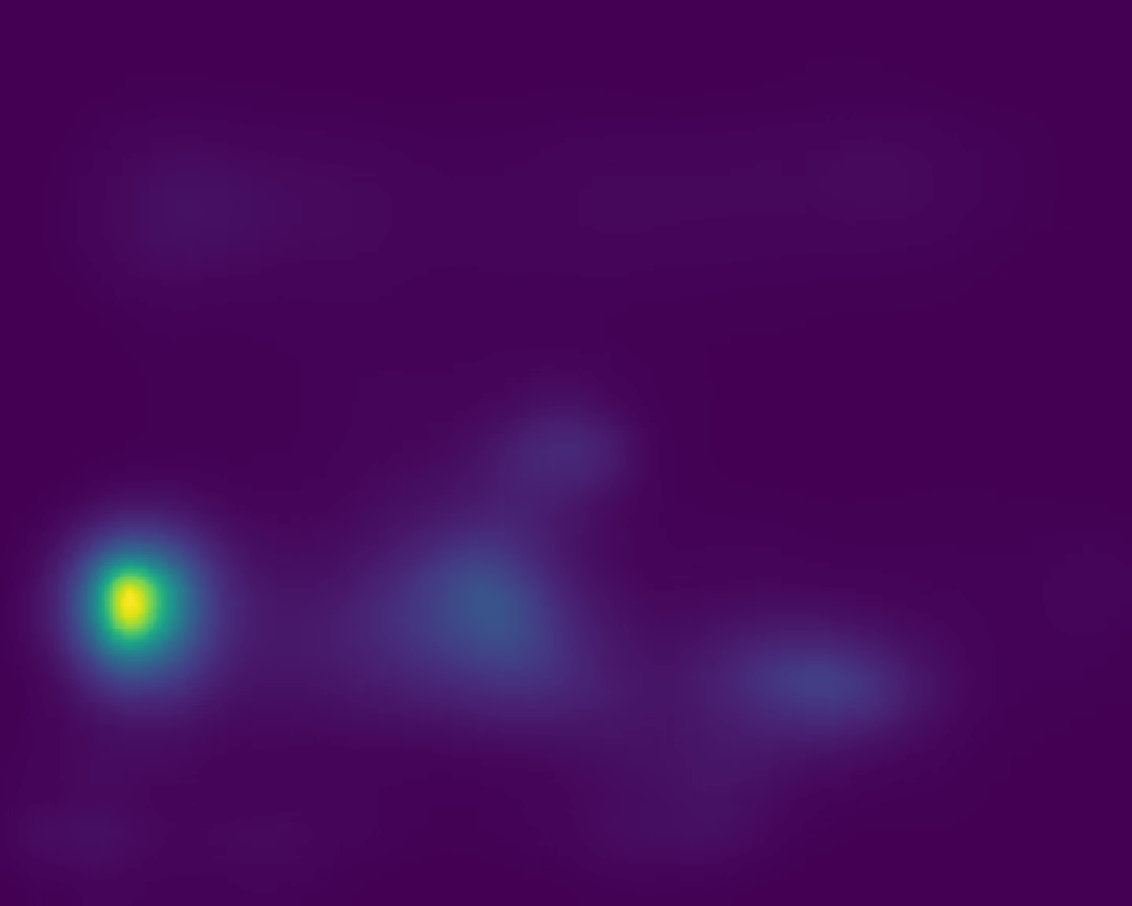}
\captionsetup{labelformat=empty}
\caption{SAM}
\end{subfigure}
\begin{subfigure}{\myspace\textwidth}
\includegraphics[width=1\linewidth]{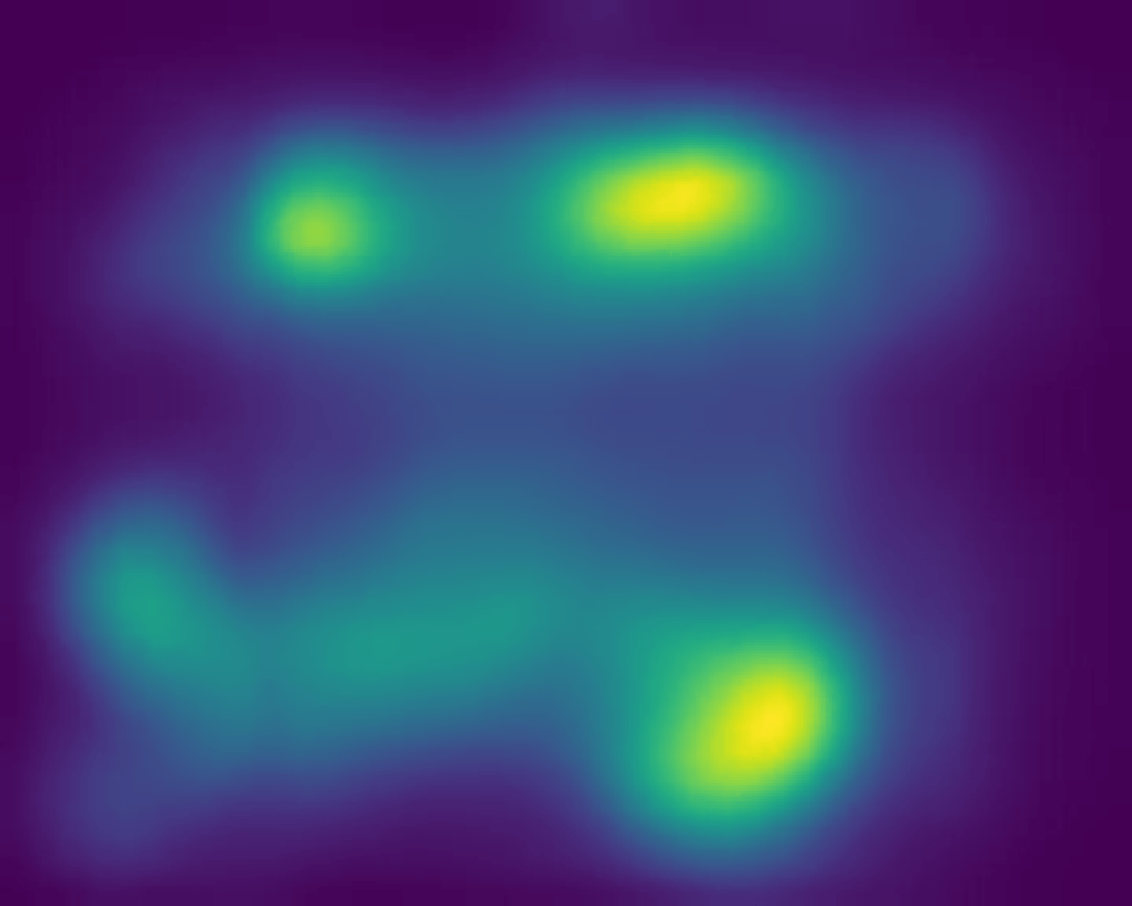}
\captionsetup{labelformat=empty}
\caption{ACL}
\end{subfigure}
 
\caption{Examples of videos frames and their empirical and generated saliency maps.}
\label{fig:saliency_data}
\end{figure}

For each type of saliency data, we personalize the baseline model in the same way as sections \ref{section:calibration} and \ref{section:quantity_test} where, for each participant, the data from the video phase is used for fine-tuning and data from the point phase is used for testing. Essentially, the network is seeing the same inputs but with different loss map labels. Quantitative results are shown in Table \ref{table:generated_saliency}. We can see that in terms of performance, Empirical $>$ ACL $>$ SAM. This is consistent with the assumption that empirical saliency from eye trackers is more accurate than their generated counterpart. Furthermore, since we use video content for data collection, it makes sense that using the video-based algorithm ACL performs better than the image-based algorithm SAM. Nevertheless, using any of the three types of saliency data show clear improvement over the baseline.

\begin{table}[ht!]
\centering
\begin{tabular}{l|c|c} 
\hline
Saliency Type & Error (cm) & Improvement (\%) \\
\hline\hline
Baseline & 4.38 $\pm$ 1.18 & 0 \\ 
\hline
Empirical & 3.32 $\pm$ 0.77 & 24.20 \\
\hline
SAM & 3.99 $\pm$ 0.58 & 8.90 \\
\hline
ACL & 3.70 $\pm$ 0.54 & 15.53 \\
\hline
\end{tabular}
\caption{Gaze estimation error using different types of saliency computations.}
\label{table:generated_saliency}
\end{table}

%------------------------------------------------------------------------
\section{Discussion}
\label{section:discussion}
%------------------------------------------------------------------------
In this section we discuss two main challenges of using saliency data for gaze calibration.

\subsection{Imprecise Labels}
One of the biggest differences between our approach and traditional gaze estimation methods is that we do not have precise ground truth gaze labels. Since there is no constraint on where the user is looking, there is a portion of data where the input image is not consistent with the associated saliency map. This portion of data can be interpreted as mislabelled data which will hinder the network from learning the correct parameters. In this paper, we tried to address this issue using an iterative outlier removal technique. However, we believe that using a more sophisticated method to filter the training data such that there is higher correlation between the input images and saliency labels as well as selecting genres of videos with clearer saliency regions (e.g., sports) will further improve the performance.

\subsection{Unbalanced Data}
\begin{figure}[h]
\centering

\rotatebox[origin=c]{90}{Coutrot 1}
\begin{subfigure}{0.14\textwidth}
\includegraphics[width=1\linewidth]{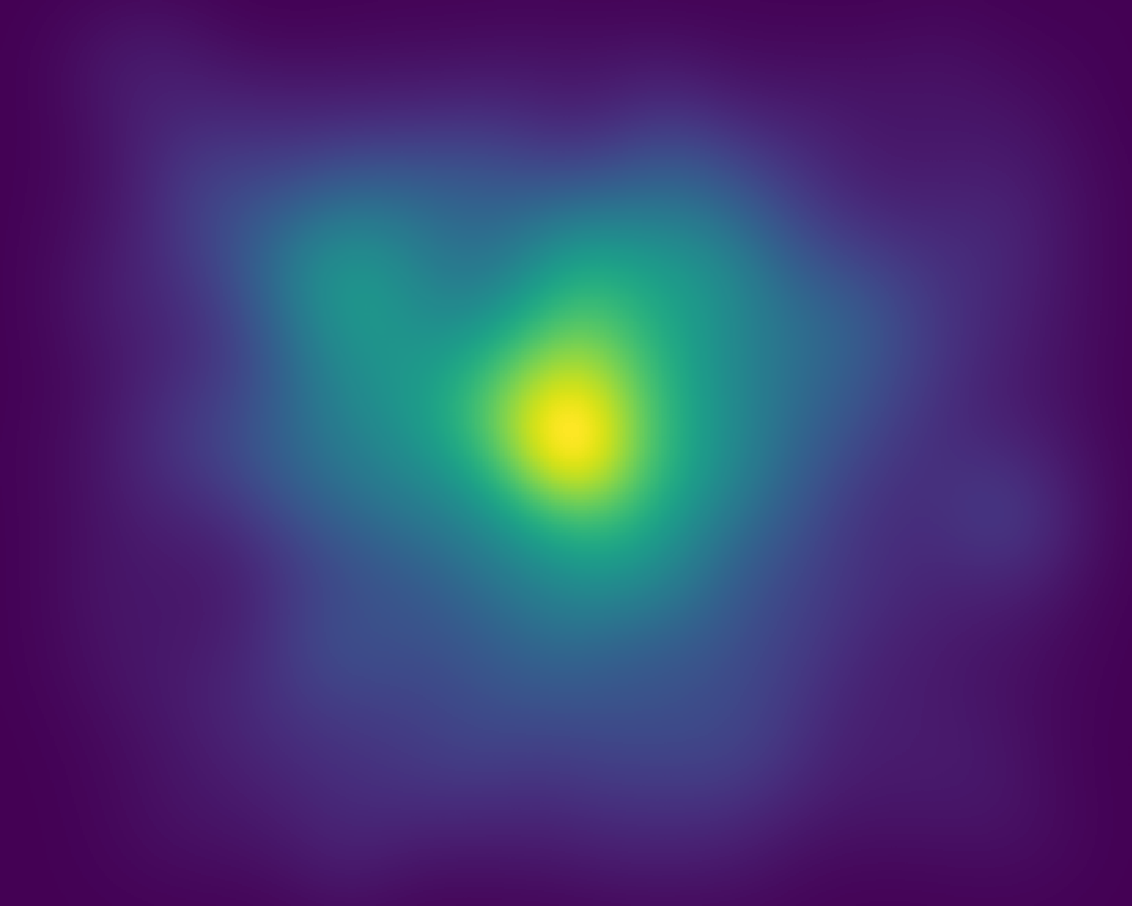}
\end{subfigure}
\begin{subfigure}{0.14\textwidth}
\includegraphics[width=1\linewidth]{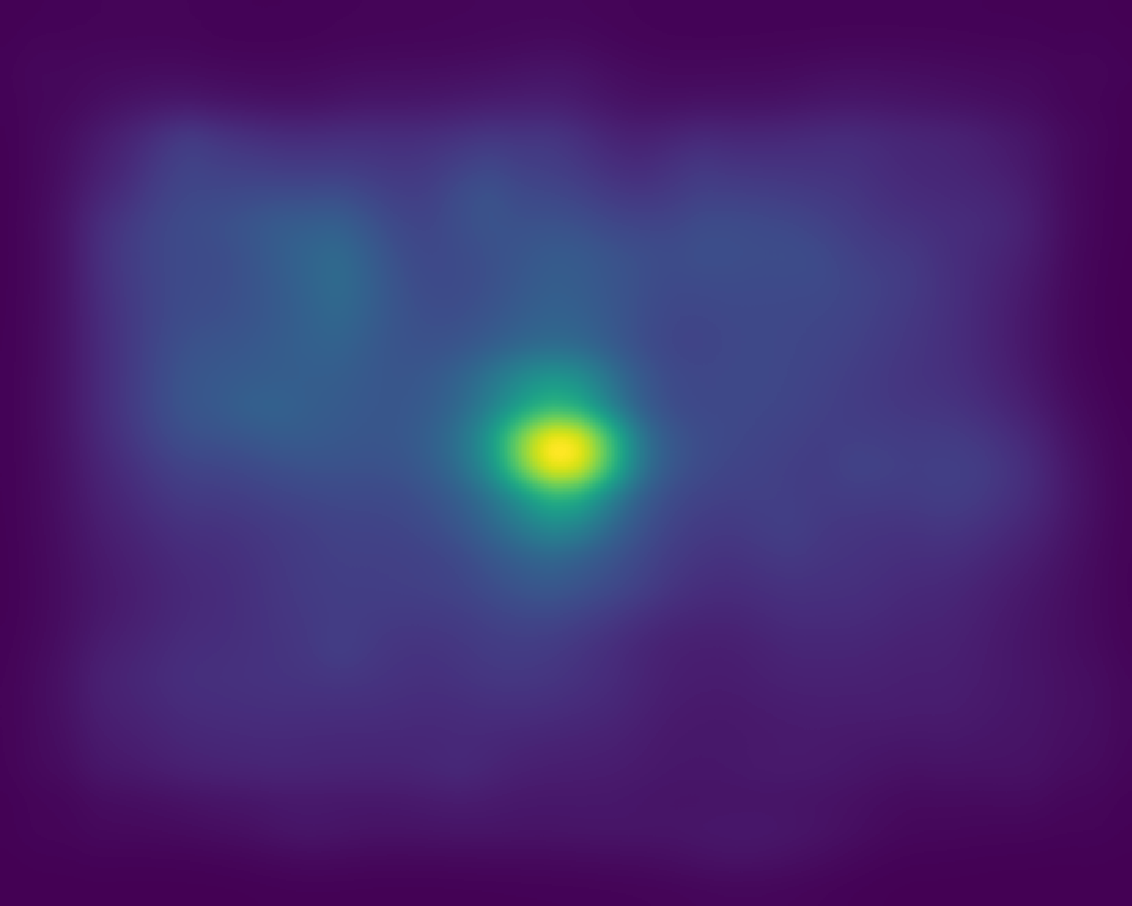}
\end{subfigure}
\begin{subfigure}{0.14\textwidth}
\includegraphics[width=1\linewidth]{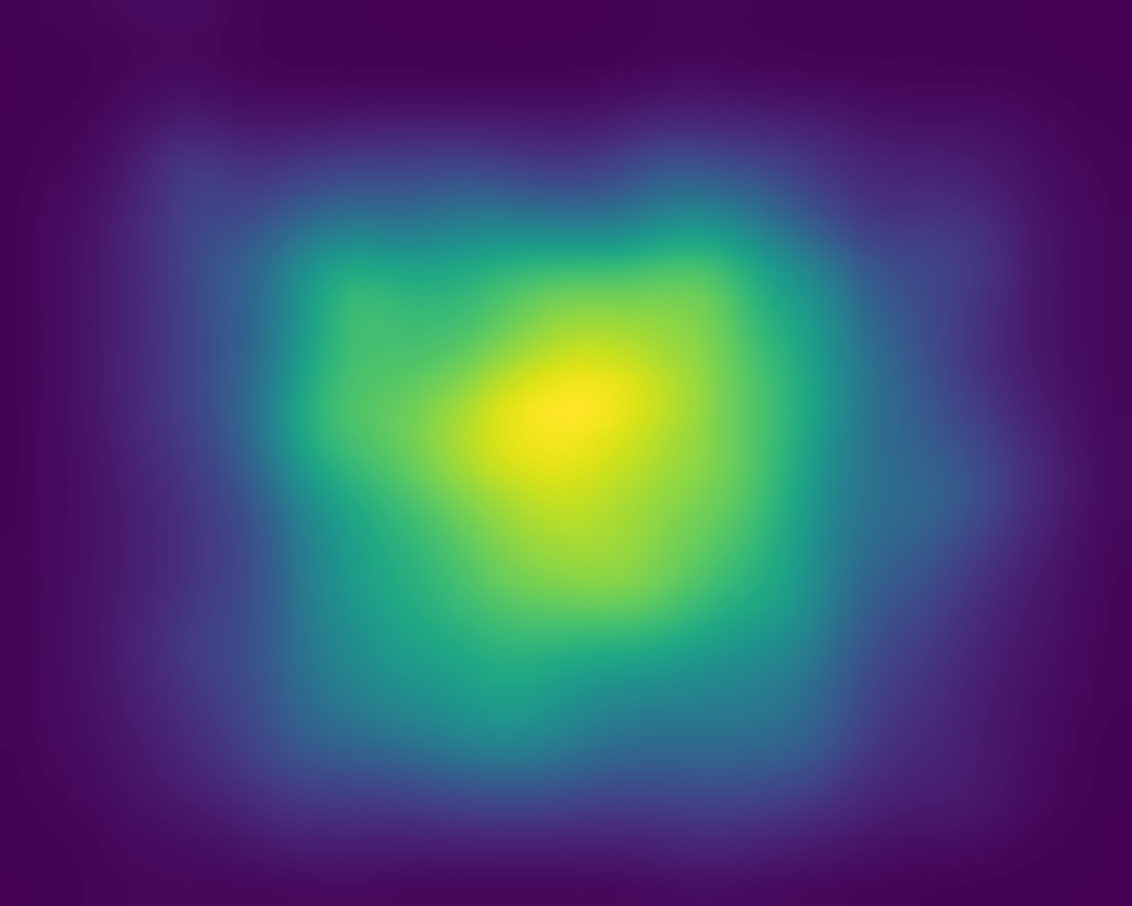}
\end{subfigure}

\rotatebox[origin=c]{90}{\hspace{11pt} SAVAM}
\begin{subfigure}{0.14\textwidth}
\includegraphics[width=1\linewidth]{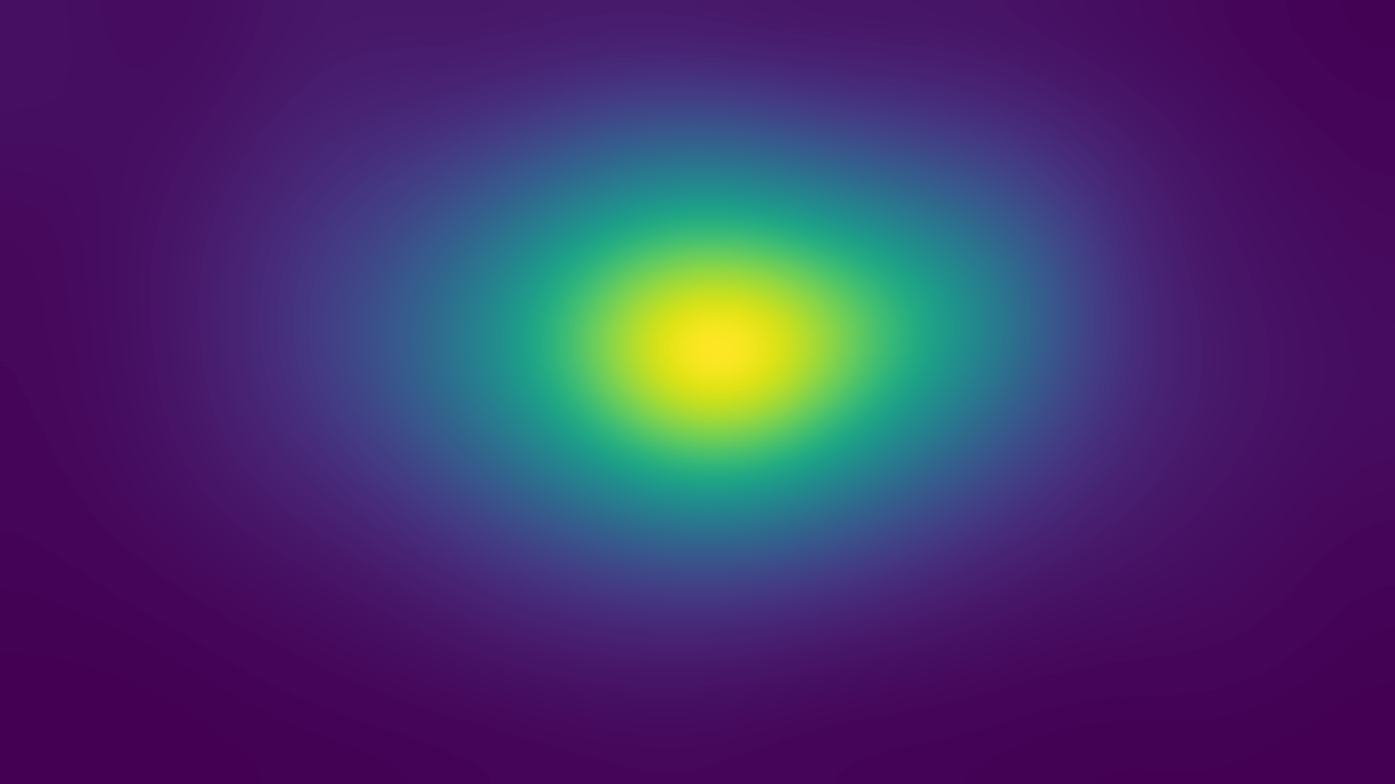}
\captionsetup{labelformat=empty}
\caption{Empirical}
\end{subfigure}
\begin{subfigure}{0.14\textwidth}
\includegraphics[width=1\linewidth]{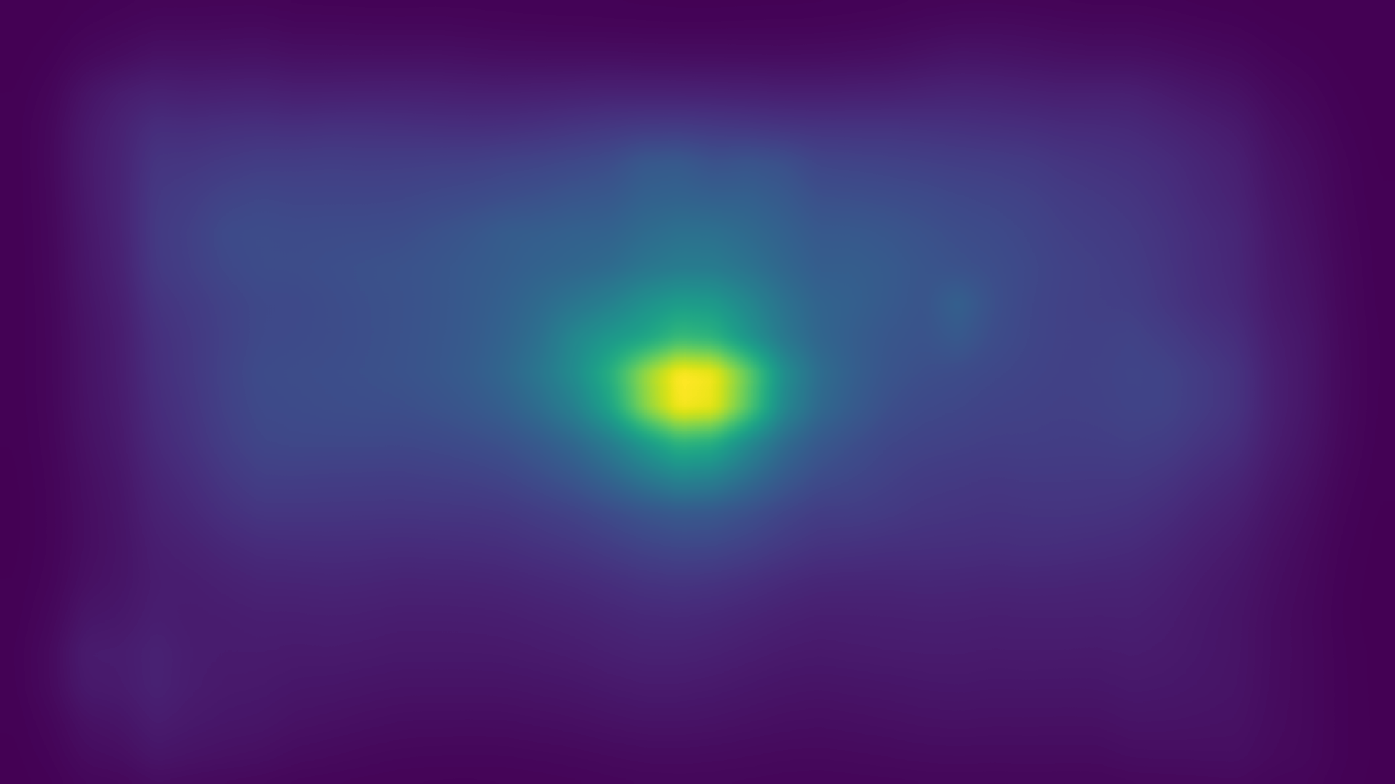}
\captionsetup{labelformat=empty}
\caption{SAM}
\end{subfigure}
\begin{subfigure}{0.14\textwidth}
\includegraphics[width=1\linewidth]{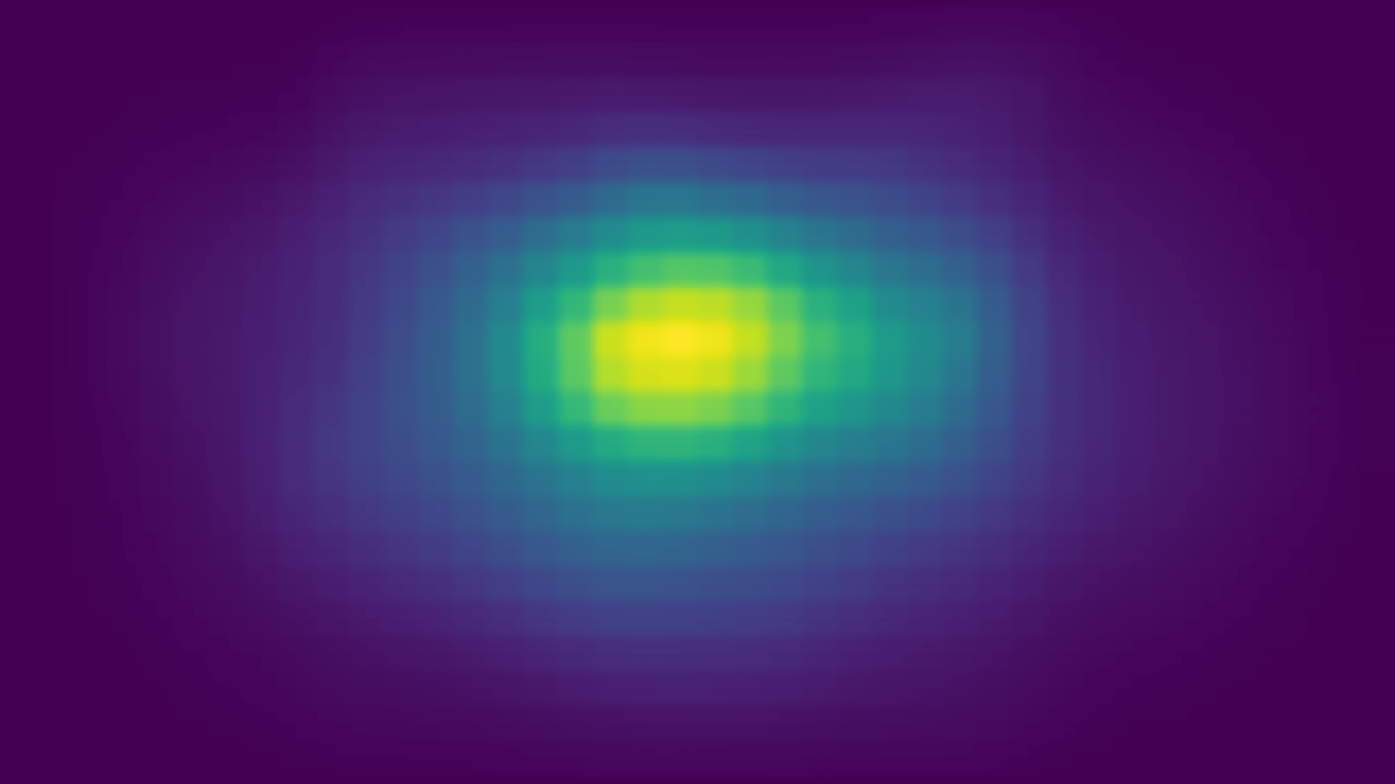}
\captionsetup{labelformat=empty}
\caption{ACL}
\end{subfigure}
 
\caption{Average saliency distributions of all videos from SAVAM and Coutrot Database 1.}
\label{fig:average_saliency}
\end{figure}

In the study of visual saliency, the center bias is a well known phenomenon where objects of interest appear more often near the center of the scene \cite{tseng2009quantifying, tatler2007central}. Figure \ref{fig:average_saliency} shows the average distribution of the empirical and generated saliency maps across all videos for SAVAM and Coutrot Database 1. This non-uniformity skews the prediction to favor the center region which results in larger errors at the boundaries. We think one area of future work could be directed at correcting or compensating for this bias, e.g., sample the training data to have a more uniform distribution, or use a spatially weighted loss. 

%------------------------------------------------------------------------
\section{Conclusion}
%-------------------------------------------------------------------------
In this paper, we proposed SalGaze, a novel framework for gaze estimation using visual saliency information. We designed an algorithm to transform a saliency map into a differentiable loss map that is well suited for the optimization of CNN-based models. SalGaze is able to combine implicit video calibration data with explicit point calibration data using a unified framework. Our technique does not require explicit attention from the user and can run in the background while the user uses the device. This lack of constraints may lead to outliers in the training data. We show that we are able to partially overcome this issue by means of an iterative outlier removal procedure. Our method also enables the collection of large amounts of gaze data which is critical for deep learning based algorithms. We show accuracy improvements over 24\% after adapting a state-of-the-art gaze estimation algorithm with saliency information using SalGaze. 

% Three main properties make our approach robust and stable: the derived loss map is smooth and can be transparently optimized using both saliency and point gaze information. The values of the loss map have a geometrical meaning as they represent the euclidian distance to a set that (most likely) contain the ground truth gaze. The gradient of the loss map is well defined and has constant (unitary) magnitude which leads to a stable and robust gradient descent optimization. 

% Our technique does not require the cooperation of the user and can run in the background while the user uses a device without his/her explicit interaction. This lack of constraints may lead to outliers in the training data. We show that we are able to partially overcome this issue by means of an iterative outlier removal procedure.

\section*{Acknowledgements}
%-------------------------------------------------------------------------
Work partially supported by NSF, ARO, ONR, NGA, NIH, Simons Foundation, Amazon AWS, and Microsoft. 

{\small
\bibliographystyle{ieee}
\bibliography{egbib}
}

\end{document}